\pdfoutput=1
\documentclass[10pt,a4paper]{article}

\usepackage[square,sort,comma,numbers]{natbib}
\usepackage{amsmath}
\usepackage{amsfonts}
\usepackage{amssymb}
\usepackage{amsthm}
\usepackage{verbatim}
\usepackage{enumerate}
\usepackage{natbib}
\usepackage{mdwlist} % for compact lists
\usepackage{color}
\usepackage[utf8]{inputenc}

\usepackage{algpseudocode}
\usepackage{algorithmicx}
\usepackage{lineno}
\usepackage{framed,multirow}
\usepackage[final]{pdfpages}
\usepackage{latexsym}
\DeclareMathOperator*{\argmax}{arg\,max}
\usepackage{url}
\usepackage{xcolor}
\usepackage{graphicx, subfigure}
\usepackage{verbatim}

\RequirePackage[left=2cm,%
                right=2cm,%
                top=2.25cm,%
                bottom=2.25cm,%
                headheight=12pt,%
                letterpaper]{geometry}%

\theoremstyle{definition}

\title{Multilayer Complex Network Descriptors for Color-Texture Characterization}

\author{Leonardo F S Scabini$^{1}$, Rayner H M Condori$^{2}$, Wesley N Gonçalves$^{3}$, Odemir M Bruno$^{1,2}$}
\date{}
\begin{document}
\maketitle
\noindent{$^1$São Carlos Institute of Physics, University of S\~{a}o Paulo, S\~ao Carlos - SP, 13566-590, Brazil.}

\noindent{$^2$Institute of Mathematics and Computer Science, University of S\~ao Paulo, S\~ao Carlos - SP, 13560-970, Brazil.}

\noindent{$^3${Federal University of Mato Grosso do Sul, Ponta Por\~{a},  Mato Grosso do Sul, Brazil}

\begin{abstract}
A new method based on complex networks is proposed for color-texture analysis. The proposal consists on modeling the image as a multilayer complex network where each color channel is a layer, and each pixel (in each color channel) is represented as a network vertex. The network dynamic evolution is accessed using a set of modeling parameters (radii and thresholds), and new characterization techniques are introduced to capt information regarding within and between color channel spatial interaction. An automatic and adaptive approach for threshold selection is also proposed. We conduct classification experiments on 5 well-known datasets: Vistex, Usptex, Outex13, CURet and MBT. Results among various literature methods are compared, including deep convolutional neural networks with pre-trained architectures. The proposed method presented the highest overall performance over the 5 datasets, with 97.7 of mean accuracy against 97.0 achieved by the ResNet convolutional neural network with 50 layers.
\end{abstract}
\section{Introduction}

Texture is a unique physical property found on most natural surfaces. We can discriminate materials, animals, and plants by observing texture patterns, which makes texture analysis an essential skill of our visual system. On a digital image, these characteristics are capt by patterns over the pixel distribution. However, there isn't a formal definition of the term texture that is widely accepted by the scientific community. One can consider the definition of Julesz \cite{julesz1962visual,julesz1973inability}, which consider two similar textures if their first and second orders statistics are similar. In a simple fashion, we can consider texture as a combination of intensity constancy and/or variations over the pixel distribution, forming a spatial pattern. Although, the physical world imposes on these raw intensity changes a wide variety of spatial organization, roughly independently at different scales \cite{marr1982}. The challenge of texture analysis is to identify and characterize these patterns to discriminate between images. Therefore, there is a constant need for new methods for texture characterization on areas such as industrial inspection \cite{industrial2003}, remote sensing \cite{remotesensing}, biology \cite{taxonomy2011, casanova, wesley}, medicine \cite{medical}, nanotechnology \cite{wesley}, biometry \cite{goncalves2010}, etc.

Most of texture analysis methods only consider the image gray-levels to characterize it. The color information is usually ignored, or processed separately with non-spatial approaches such as color statistics (histograms, statistical moments, etc). Studies have been carried out showing the benefits of adding color for texture analysis \cite{DRIMBAREAN20011161, maenpaa2004, bianconi2011theoretical,cernadas2017}. However, most methods of color-texture analysis do not consider the spatial relation of pixels in different color channels.

% Recently, Convolutional Neural Networks (deep learning) are attracting a lot of attention from the CV community for image characterization. After the results of the large neural network AlexNet \cite{krizhevsky2012imagenet}, many thought this approach was the definitive solution for pattern recognition. Then several convolutional architectures were proposed, or authors simply apply a pre-trained network, achieving state-of-the-art results in various scenarios. However, due to the lack of thorough understanding of the limitations of existing neural architectures \cite{basu2016theoretical}, some intriguing properties was observed on convolutional networks \cite{szegedy2013intriguing, nguyen2015deep}. Experiments have shown that some imperceptible changes in an image can cause the network to label it as something else entirely. It is possible to produce images that are completely unrecognizable to humans, but that convolutional networks believe to be recognizable objects \cite{nguyen2015deep}. We then question whether they are applicable for texture analysis.

In this work, we propose Multilayer CN descriptors, a new color-texture analysis which capt within and between color-channel spatial interaction by modeling the full-color image with a multilayer CN. In our proposal, each color-channel is a network layer, and pixels within-between color channels can be connected. We also use an automatic and adaptive threshold selection technique at the modeling step, to reduce the number of parameters and improve the method performance. The structure of the resulting multilayer CN is split into within and between layer connections, which allows a better analysis of color-texture. Moreover, we propose a set of statistical measures from the degree and clustering distributions to characterize the CN topology, composing a color-texture descriptor. We show through classification experiments in 5 datasets that the proposed method outperformed traditional and recent color-texture descriptors, and also some well known convolutional neural networks with pre-trained architectures.

The work follows divided into the sections: Section \ref{sec:TC} presents and review on texture analysis methods and the mathematical formulation of CNs; Section \ref{sec:PA} describes the proposed approach, Section \ref{sec:experiments} shows all the experiments and its results, comparisons with literature methods and discussions; finally, on Section \ref{sec:conclusion} we discuss the main findings and results of the work.

\section{Theoretical Concepts and Review}\label{sec:TC}
On this section, we present the theoretical concepts of color-texture and CN.

\subsection{Color-Texture Analysis}

The goal of texture analysis is to study how to measure texture aspects and employ it to image characterization. Although this approach has been explored for many years, most works approach gray-level images, i.e. a single color-channel representing the pixel luminance. On this case, the multispectral information is either discarded or is not present. However, nowadays images that derive from different sources are mostly colored, for instance from the internet, surveillance cameras, satellites, microscopes, personal cameras and much more. The recent increase in computer hardware performance also makes possible to analyze larger amounts of data, allowing to keep the color information for texture analysis.

In general, color-texture methods found in the literature are mostly integrative, which separate color from texture. These methods usually compute traditional gray-level descriptors from each color-channel, separately. For that, various gray-level methods can be applied or combined into an integrative descriptor. There are many gray-level methods found in literature. Classical techniques can be divided into statistical, model-based and structural methods \cite{methods}. The statistical methods were one of the first approaches which considered texture as a property to characterize images. Among statistical methods, the most common ones are those based on gray-level co-occurrence matrices \cite{haralick, PALM2004integrativeCooccurrence} and Local Binary Patterns (LBP) \cite{nanni2012survey}. Model-based methods include descriptors such as Gabor filters \cite{gabor}, which explore texture in the frequency domain, and Markov random field models \cite{panjwani1995markov}. Structural methods consists on analyzing the texture as a combination of various smaller elements, that are spatially arranged to compose the overall texture pattern. This is achieved, for instance, through morphological decompositions \cite{lam1997rotated}. More recent and innovative techniques are approaching texture differently. For instance, there are texture descriptors based on image complexity analysis with fractal dimension \cite{backes2008fractal,fractal}, Complex Networks (CNs) \cite{backes2012,wesleydinamic,scabini2015texture,xu2015complex,gonccalves2016texture} and Cellular Automaton \cite{CA2015texture}. Learning techniques are also gaining attention, for instance texture properties learned with a vocabulary of Scale Invariant Feature Transform (SIFT) \cite{densesift}, or using Convolutional Neural Networks (CNN) \cite{cimpoi2015deep}.

%A pioneer work was proposed by Haralick and Karthikeyan \cite{haralick}. They measure the texture by computing gray-level co-occurrence matrices of pixels intensity. The patterns of co-occurrence are quantified with statistical measures of signal processing. Then many works inspired on co-occurrence matrices was proposed  \cite{davis1979texture, gotlieb1990texture, argenti1990fast}. Another important work was proposed by Ojala et al. \cite{ojala1996comparative, ojala2002multiresolution} with the statistical method named Local Binary Patterns (LBP). An LBP is obtained by thresholding a neighborhood with the value of the center pixel and considering only the sign information. This motivated several later works and LBP-based methods continue to emerge (the reader may consult \cite{nanni2012survey} for a survey of LBP-based methods). 

Pure color methods are another kind of approach for color-texture analysis which considers only the image colors. On this class of methods, the most diffused are those based on color histograms \cite{hafner1995efficient}. This technique provides a compact summarization of the color distribution. Usually, this is done by first discretizing the color values into a smaller number of bins. Then, different approaches can be applied to count the occurrence of pixels. For instance, joint histograms can be obtained by combining pairs of color channels, or also 3D histograms, considering all channels. However, these methods do not consider the pixel spatial interaction, as gray-level texture methods do. Therefore, a common approach is to combine pure color and gray-level features into parallel methods. On \cite{maenpaa2004, cernadas2017} the reader may consult evaluations of different integrative, pure color and parallel methods. In some cases, integrative methods perform similarly to traditional gray-level features computed using only the pixel luminance, which leads some authors to question their applicability. Therefore, we believe that this is not the best approach to address color for texture analysis. Color and texture must be analyzed in a more intrinsic approach, considering relations between spatial and spectral patterns.

Considering the human visual system, having color in images helps to recognize things faster and to remember them better \cite{gegenfurtner2000,wichmann2002contributions}. Research in psychophysics \cite{zhang1997spatial} have shown that the human perception of color depends on the frequency with which the color is spatially distributed, rather than colors forming uniform areas. We can also consider the color opponent process theory \cite{foster1895text}, which suggests that the human visual system focus on variations between pairs of colors rather than each individual color. As texture underlies on patterns of intensity change, we believe that rich color-texture information can be extracted by analyzing the spatial interaction between image color-channels. One of the first works to address within and between color-channel interaction to texture analysis was proposed by Rosenfeld et. al. \cite{rosenfeld1980multispectral}. The authors have explored the interactions between pairs of color-channels (that they called two-band features), and results indicate that it provides textural information that is not available from the single-channel analysis (single-band features). This motivated further research, then various works approaching color-channel relations have been proposed. For instance, in \cite{mirmehdi2000segmentation} color-texture is considered for image segmentation by approaching the interplay of colors and their spatial distribution in an inseparable way as they are actually perceived during the preattentive stage of human color vision. In \cite{casanova2016}, the authors propose the use of fractal descriptors estimated by the mutual interference of color channels.

%TODO CONTINUAR REVISAO DE METODOS WITHIN-BETWEEN COLOR CHANNEL

%We also must consider that few models explore pure color features to represent the texture. Therefore, we believe that in texture analysis, color is a poorly explored attribute that needs further investigation.

%The goal is to propose models that present some coherence between performance, number of features and computational complexity. Moreover, it is also desirable that the method presents tolerance to transformations such as rotation and scale changes, and the influence of noise.
%\subsubsection{Traditional methods}

%Recently, some authors \cite{tivive2006texture, cimpoi2014describing,cimpoi2015deep,cimpoi2016,basu2016theoretical, Liu2016} have studied the performance of Convolutional Neural Networks (deep learning) on texture classification. However, one of the issues with these methods is the lack of thorough understanding of the limitations of existing neural architectures \cite{basu2016theoretical}. Moreover, some traditional texture descriptors have a better overall performance than deep learning methods when the computational complexity is taken into consideration \cite{Liu2016}, which limits their application for real-time systems.

\subsection{Complex Networks}

The CN research arises from the combination of the graph theory, physics, statistical mechanics and computer science, with the goal of analyzing large networks derived from complex natural (real-world) processes. Initially, works have shown that some structural pattern is present in most of these networks, something that is not expected in a random network. This led to the definition of CN models that allows the understanding of its structural properties. The most popular ones are the scale-free \cite{scalefreeCN} and the small-world model \cite{smallworldCN}. Therefore, a new door was opened for pattern recognition, where CN is employed as a tool for modeling and characterization of natural phenomena.

We can observe the application of CN concepts in various areas such as physics \cite{kitsak2010identification}, nanotechnology \cite{machado2017complex}, neuroscience \cite{rubinov2010complex}, and many more \cite{aplicacoesRC}. This is an emerging science which is gaining strength due to big data and recent faster computer hardware, that enables the processing of larger amounts of data. The incorporation of CNs concepts in some problem consists of two principal steps: i) the modeling of the system and; ii) the analysis of the resulting network. By quantifying the CN topology we can draw important conclusions related to the system it represents. Local vertex measures can highlight important regions of the network, estimate its vulnerability, find groups or clusters of similar vertices, etc.

Basically, a network $N=\{V, E\}$ is composed of a set of vertices $V=\{v_1, ..., v_n\}$ and a set of edges (connections) $E=\{w(v_i, v_j)\}$, where $w$ can be some metric (weighted networks) or simply binary, indicating if $v_i$ and $v_j$ are connected or not (unweighted networks). The edge can also have a direction (directed networks), indicating if $v_i$ points to $v_j$, or the opposite. We consider only undirected weighted networks on this work. The CN topology is defined by its pattern of connections. To quantify these patterns, some measures can be extracted whether for a specific vertex, a group of vertices or the entire network. Two common measures are the degree $k$ and the clustering coefficient $c$ of a vertex. The degree is the number of edges that connect a vertex $v_i$ with other vertices

\begin{equation}\label{eq:degree}
k(v_{i}) = \sum_{\forall v_j \in V}\left\{
\begin{array}{l}
1 $, \ if $w(v_i, v_j) \in E
\\
0 $,  \ otherwise$
\end{array}
\right
.
\end{equation}

The degree is a measure of how the vertex interacts with its neighbors. The degree distribution can help to draw important conclusions about the network structure. For instance, the scale-free model \cite{scalefreeCN} is known for having a power law degree distribution of type $P(k) = k^{-\gamma}$ ($P(k)$ is a probability density function). In other words, these networks have few clusters, which are vertices densely connected, and many vertices with few connections. Many findings indicate that most real-world networks have $\gamma \approx 3$.

The clustering coefficient of a vertex is another way to measure neighborhood interplay on the network. Different from the degree, which only counts the number of connections of a vertex, the clustering considers connections between the vertex neighbors as well. It basically measures the structure of connections of a group of vertices by counting the number of triangles that occur between them. A triangle occurs whenever a triple of vertices is fully connected. Consider $N_i$ as the neighbors directly connected to vertex $v_i$. The clustering coefficient of $v_i$, considering that $N$ is undirected, is defined by the fraction

\begin{equation}\label{eq:clustering}
c(v_i) = \frac{2 |w(v_j, v_k) : v_j, v_k \in N_i, w(v_j, v_k) \in E|}{k(v_i)(k(v_i)-1)}        
\end{equation}
where the numerator is the number of edges between the neighbors of $v_i$ multiplied by 2. The equation is normalized by the maximum number of possible triangles $v_i$ can form with its neighbors ($k(v_i)(k(v_i)-1)$). 

The clustering coefficient is helpful to understand the small-world property \cite{smallworldCN}, a phenomenon that occurs when vertices can be reached through a relatively small path. On these networks, the average geodesic path length (average distance between all vertices) is low, while the average clustering coefficient is high.

\subsection{Complex Networks on Texture Analysis}

The CN theory is also gaining attention from the CV community. From the proposal of modeling an image as a network, many possibilities arise, where the solution became a network problem. For instance, recent works are approaching image segmentation as a community detection problem \cite{browet2011community,hu2012modularity}. Specifically, on texture characterization, that is the main idea of our work, the idea arise to model images as a networks by representing each pixel as a vertex \cite{chalumeau2006}. Pairs of vertices are connected with weight equivalent to its absolute intensity difference. Connections are then transformed addressing 2 parameters, a radius for spatial distance limiting and a threshold for connection weight pooling, where high-weighted connections are removed. This approach was then introduced for boundary shape analysis \cite{backes2009complex}, where a network is obtained by connecting all contour pixels. On this case, the euclidean distance between vertices is considered as the connection weight, and CNs are obtained by thresholding, keeping connections between spatially close contour pixels. On \cite{backes}, spatial information was introduced on the definition of the edge weight for texture analysis, and the dynamic evolution of the network is evaluated by a set of different thresholds.

In \cite{wesleydinamic}, the original CN approach is extended to model dynamic texture (videos) by connecting vertices/pixels from different frames. The characterization is then made by vertex measures considering connections in the same frame or between subsequent frames, allowing both spatial and temporal analysis. In \cite{scabini2015texture}, the CN approach is combined with Bag-of-visual-words to build a visual vocabulary, defined in a training step to cluster image regions based on network local descriptors. In the most recent work \cite{gonccalves2016texture}, the authors explore the concept of diffusion and random walks on CNs modeled from texture images. All these works focus only on gray-level images, i.e. the texture is represented by a 2D matrix where pixel values are the luminance.

In \cite{sajunior2014} the authors propose a technique to model color-texture images as graphs and then compute shortest paths between vertices. Although the approach is similar to CN works, the modeling step only considers the distance between vertices to create connections. In other words, the structure of the resulting network is regular because the degree distribution is constant, i.e all vertices have the same number of connections (except border vertices). A regular network cannot be evaluated with CN techniques, as we measure the pattern of connections between vertices.% In this work, we propose the modeling of a multilayer CN which can be analyzed with topological measures in order to characterize color-texture. Classification results indicate that our method overcomes the graph-based approach \cite{sajunior2014} and other methods from the literature (see results on Table \ref{comparison}).

\section{Multilayer Complex Networks on Color-Texture}\label{sec:PA}

The main contribution of this work is the proposal of a new technique to model and characterize color-texture with CN, which we describe in the following. By addressing the concept of a multilayer CN, it is possible to map each color-channel as a network layer, where vertices in a layer represent pixels. Consider an image $I$ with width $x$, height $y$ and $z$ color-channels, where each pixel $p$ of a total of $x*y$ pixels has $z$ intensity values (e.g. in RGB $z=3$). To build a CN $N=\{V, E\}$, we consider each pixel, in each channel, as a vertex. Thus, the total number of vertices is $|V|=x*y*z=n$, composing the set $V=\{v_1,...,v_{n}\}$. In other words, the network is composed of one vertex layer for each color-channel, that combined represent the whole network structure. The rule we use to connect pixels is similar to the technique commonly used in previous works. Consider the Euclidean distance of two vertices $v_i$ and $v_j$ as $d(v_i, v_j) = \sqrt{(p(v_i, x) - p(v_j, x))^2 + (p(v_i, y) - p(v_j, y))^2}$ where function $p(v_i, x $ or $ y)$ gives the corresponding cartesian coordinate of the pixel in the image represented by the given vertex $v_i$. Note that the color-channel of the pixel is not considered as a third dimension to compute the Euclidean distance, as this does not make any geometrical or textural sense. Therefore we only use the coordinates $x$ and $y$, that relates to the position of the pixel on the image. Connections are then defined if the Euclidean distance of vertices is smaller than a limited window defined by a radius $r$. Vertices that satisfies $d(v_i, v_j)\leq r$ are connected, composing the set of edges $E=\{w(v_i, v_j)\}$. The weight of the connections between pairs of vertices $v_i$ and $v_j$ are defined by:

\begin{equation}\label{eq:connectionWeight}
w(v_i, v_j) = \Bigl( \frac{|p(v_i) - p(v_j)| + 1}{L+1} \Bigr)  \Bigl(\frac{d(v_i, v_j) +1 }{r +1}\Bigr)
\end{equation}
where $p(v_i)$ returns the intensity value of the pixel represented by $v_i$, on its corresponding color-channel. The constant $L$ is the largest intensity value possible on the image (e.g. in 8-bit images $L=255$). We use $L$ and $r$ to normalize the connection weight to values between $[0,1]$. When $d(v_i, v_j)=0$, it means that we are connecting vertices representing the same pixel but in different color-channels. On this case, the right side of the equation equals 0 and it cancels the entire equation, thus we sum 1 both on $d$ and $r$. The same is done on the left side of the equation to avoid the case when $|p(v_i) - p(v_j)| = 0$.

This process results in a $r$-scaled network $N^r$ that includes both within and between color-channel connections (Figure \ref{fig:modeling} (a) illustrates $N^r$). However, this network is regular as all vertices have the same number of connections (except border vertices). Another transformation is necessary in order to obtain a CN which provides useful topological information. Previous works \cite{chalumeau2006,backes2009complex,backes,scabini2015texture} addressed this by using a threshold $t$ to cut some connections and keeping similar vertices connected. As texture consists of the patterns of intensity variation, we propose a different approach to cut the network connections where we keep connections between distinct vertices, not similar. This highlights network connections related to regions of the image where color intensities vary. Considering our method, low connection values means high similarity. Intuitively, we keep different vertices connected by discarding low weighted connections. A new network $N^{r,t}$ is then obtained by removing connections from the set of edges $E$
\begin{equation}\label{eq:cuting}
\forall w(v_i, v_j) \in E = \left\{
\begin{array}{l}
w(v_i, v_j) $, \ if $w(v_i, v_j) > t
\\
\ \ \ \ \ \emptyset $,  \ \ \ \ \ \ \ \ \ otherwise$
\end{array}
\right
.
\end{equation}

\begin{figure}
	\centering
	\subfigure[$N^r = $ First modeling step using the raidius $r$.]{\includegraphics[angle=-90, width=0.3\linewidth]{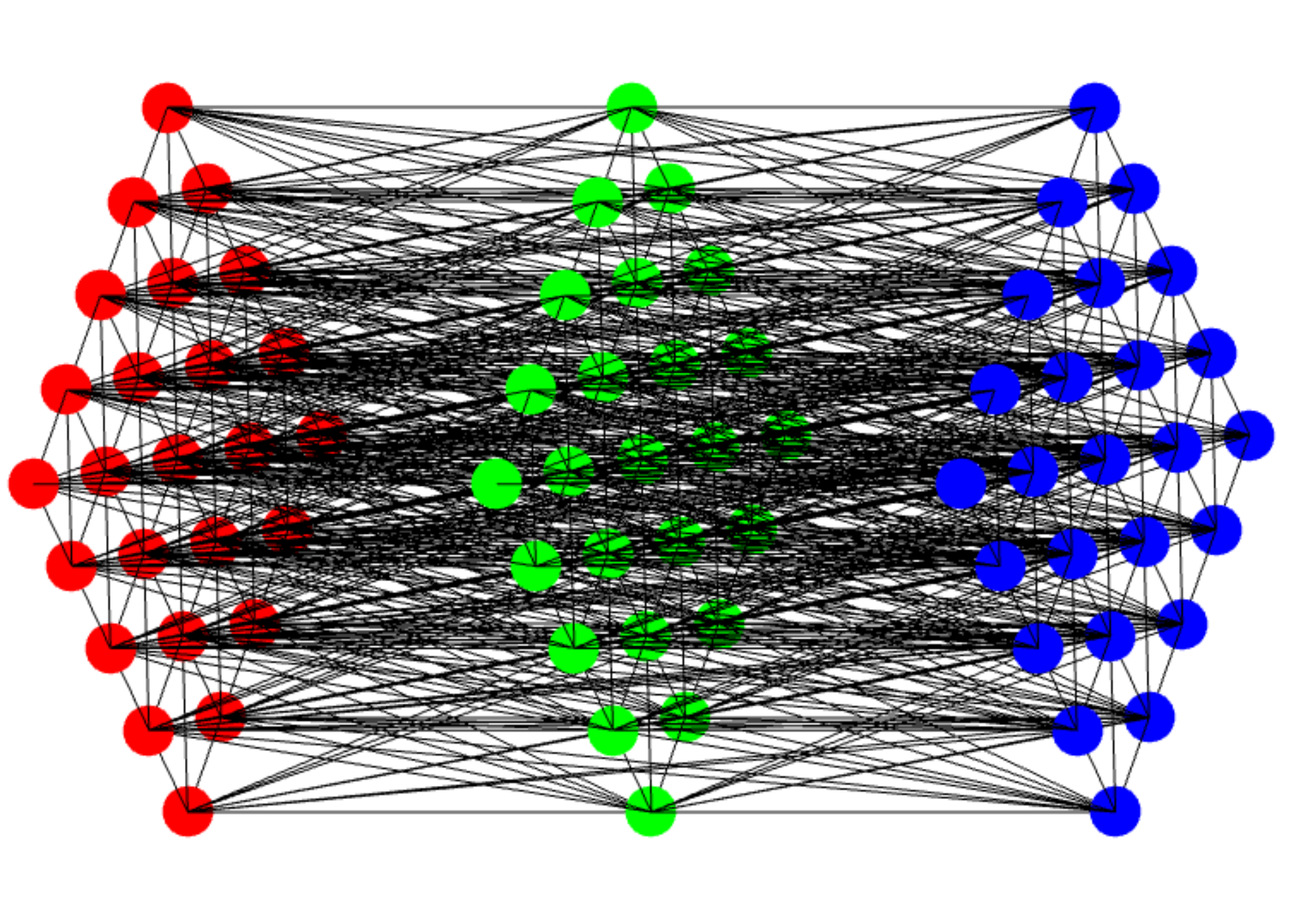}} \ \ \ \ \ \ \ \  \ \ \ \ 
	\subfigure[$N^{r,t} = $ Second modeling step where low weighted connections are removed using the threshold $t$]{\includegraphics[angle=-90, width=0.3\linewidth]{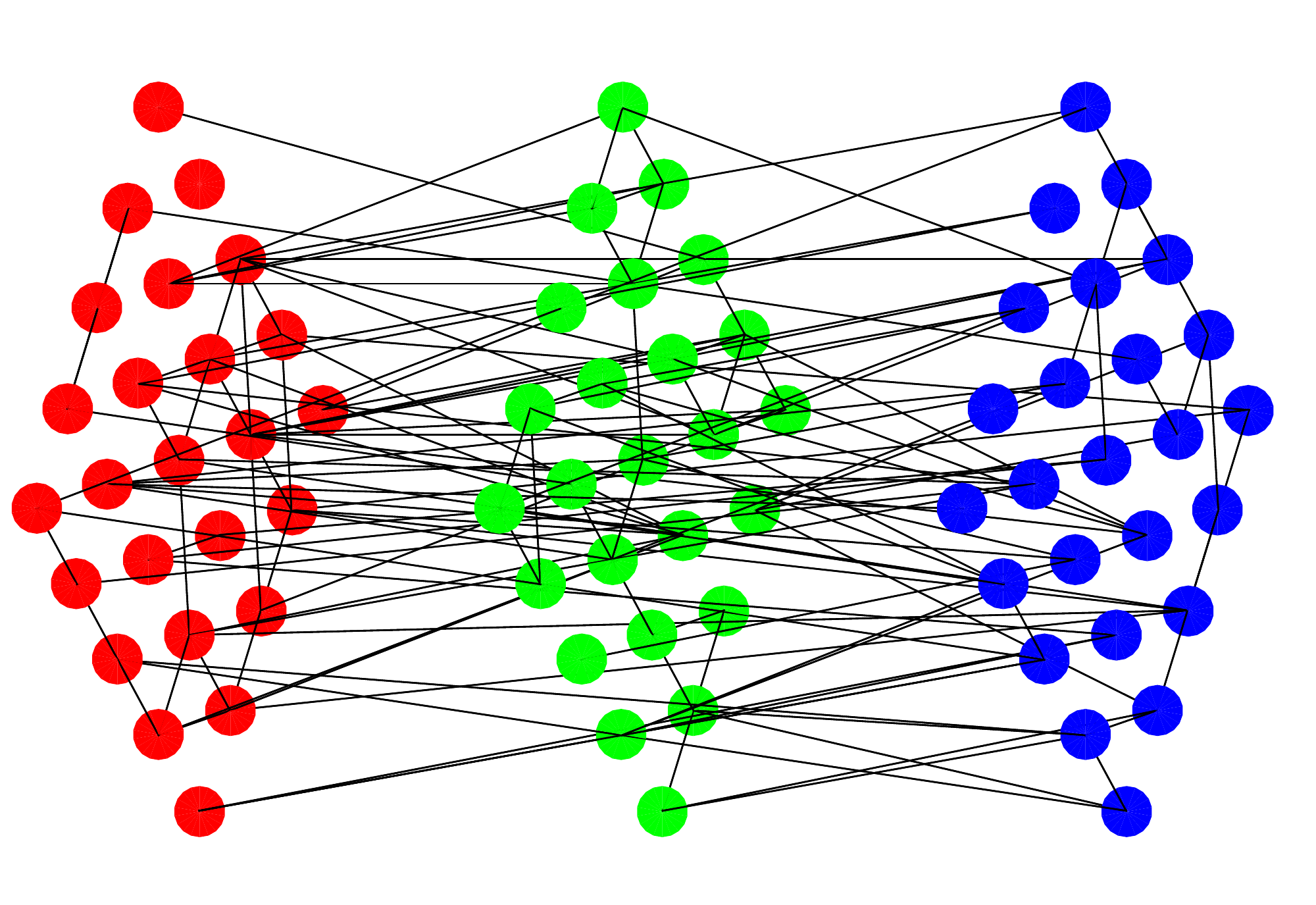}}
	\caption{\label{fig:modeling} Modeling steps of the proposed method. (a) An RGB image is modeled into a 3-layer network $N$ by representing each pixel in each channel as a vertex, and then connecting them if its Euclidean distance is smaller than a given radius $r$ (see Equation \ref{eq:connectionWeight}). (b) To obtain a complex topology useful for color-texture characterization, connections with weight smaller than a threshold $t$ are discarded (see Equation \ref{eq:cuting}).}
\end{figure}

Figure \ref{fig:modeling} (b) illustrates the new network $N^{r,t}$, which now has a topological structure that is related to the image texture. Connections capt spatial patterns of intensity variation in a specific color-channel and also the relation of spectral variation between any different channel. Thus the structure of this network contains rich information about color-texture, which can be quantified. However, the modeling parameters ($r$ and $t$) are key factors directly related to the resulting network topology. In the following section, we discuss the influence of each parameter and their benefits.

\subsection{Modeling Parameters}

The two modeling parameters of the proposed method are the radius $r$ and the threshold $t$. The value chosen for each one directly affects the resulting network structure. For instance, high radius and low threshold create dense networks, while the contrary results in sparse networks. However, we can use this topological variation to improve the network analysis. An interesting way of network characterization is to consider the interplay between structural and dynamic aspects \cite{cnusp}. Here we suggest the use of a set of radius $R=\{r_1, ..., r_i\}$ and a set of thresholds $T=\{t_1, ..., t_m\}$ to access the network dynamics. Figure \ref{fig:dynamics} shows how one layer of the network changes as $r$ and $t$ are increased.

\begin{figure}
	\centering
	\subfigure[$N^{r_2, t_1}$]{\includegraphics[width=0.2\linewidth]{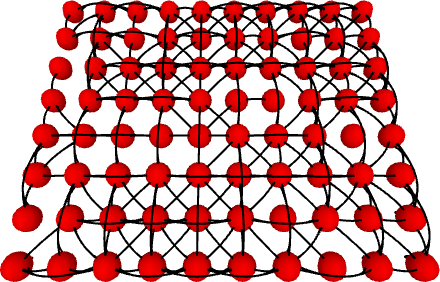}} \ \ \ \ 
	\subfigure[$N^{r_2, t_2}$]{\includegraphics[width=0.2\linewidth]{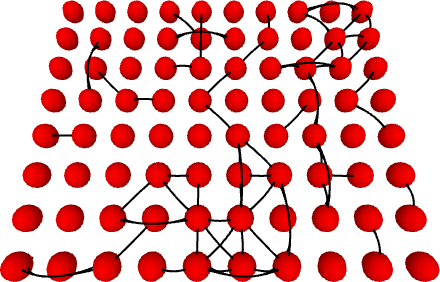}} \ \ \ \ 
	\subfigure[$N^{r_2, t_3}$]{\includegraphics[width=0.2\linewidth]{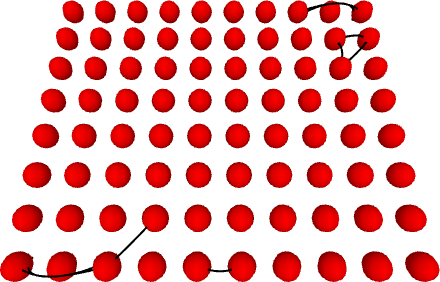}}\\
	\subfigure[$N^{r_3, t_1}$]{\includegraphics[width=0.2\linewidth]{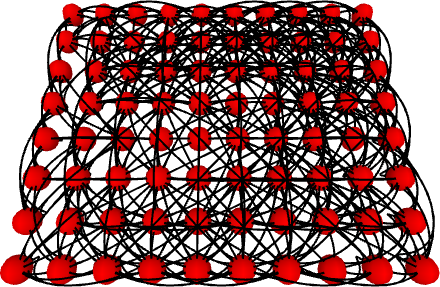}} \ \ \ \ 
	\subfigure[$N^{r_3, t_2}$]{\includegraphics[width=0.2\linewidth]{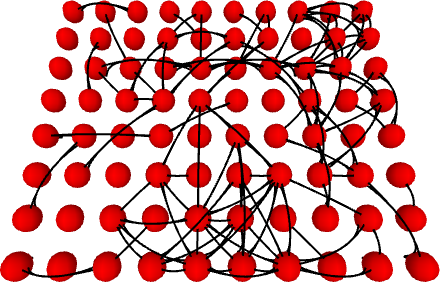}} \ \ \ \ 
	\subfigure[$N^{r_3, t_3}$]{\includegraphics[width=0.2\linewidth]{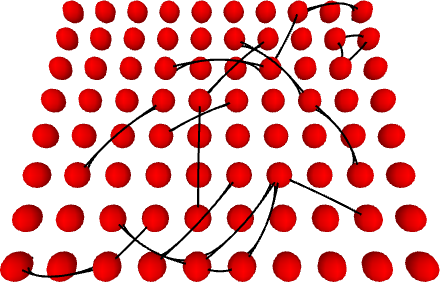}}
	
	\caption{\label{fig:dynamics} Dynamic evolution of one layer of the network $N^{r,t}$ with $r=2$ (a, b and c) and $r=3$ (d, e and f) by incrementing the threshold $t$ so that $t_1 < t_2 < t_3$.  The higher the radius and the lower the threshold, the denser the resulting network.} 
\end{figure}

To define the set $R$, we considered a circular region around the center pixel which is uniformly increased by 1 $R=\{1,2,3,4,5,6\}$. It is important to notice that one can remove the parameter $r$ and model the CN using only $t$. This will not imply in significant changes on the modeling because we already include spatial information on the edges by using the vertex Euclidean distance (see Equation \ref{eq:connectionWeight}). However, to remove the radius $r$ it is necessary to compute the connections between all possible pairs of vertices before cutting them with $t$. In most cases, the number of vertices of a color image with sizes $x*y$, given by $n=x*y*z$, is much higher than the neighborhood $r_{neigh}$ that $r$ covers ($n \gg r_{neigh}$). The removal of $r$ then imply in a relevant increase on the computational cost, from $n*r_{neigh}$ to $n^2$. Thus, the use of the radius $r$ is a way to reduce the complexity of the method considerably when dealing with common or big-sized images. On the other hand, if one needs fewer parameters over efficiency or if the image size is relatively small, this parameter can be omitted by computing the connections between all vertices and normalizing them with the highest possible value.

To define the threshold set $T=\{t_1, ..., t_m\}$ we use a similar approach to \cite{scabini2015texture,gonccalves2016texture}, that we discuss in the following.

\subsubsection{Automatic threshold selection}\label{sec:automatict}

Previous works with CN in gray-level texture \cite{backes,gonccalves2016texture} defines the threshold set empirically, by extensively testing different values. For that, 3 parameters are evaluated: an initial threshold $t_1$, an increment factor and a final threshold $t_m$. However, the edge weights of the network depend on the image colors and variations. Figure \ref{fig:edgesdistribution} shows the edge weight distribution for $N^{r_1}$ networks built for images of three color-texture datasets (See Section \ref{datasets} for detailed information). We use a discretization of 256 bins and $P(w)$ (red) is the probability of edges with weight $w$ to occur on the dataset. On the same plot we also show the mean degree $\mu_k$ (see Equation \ref{eq:degree} and Table \ref{tab:statistics}) of networks of all the dataset using the value at the x-axis as threshold ($N^{r_1, t=w}$). The range of the values of each curve is described on its corresponding y-axis (left for the edge weight distribution and right for the mean degree).

\begin{figure}
	\centering
	\subfigure[Vistex dataset]{\includegraphics[width=0.33\linewidth]{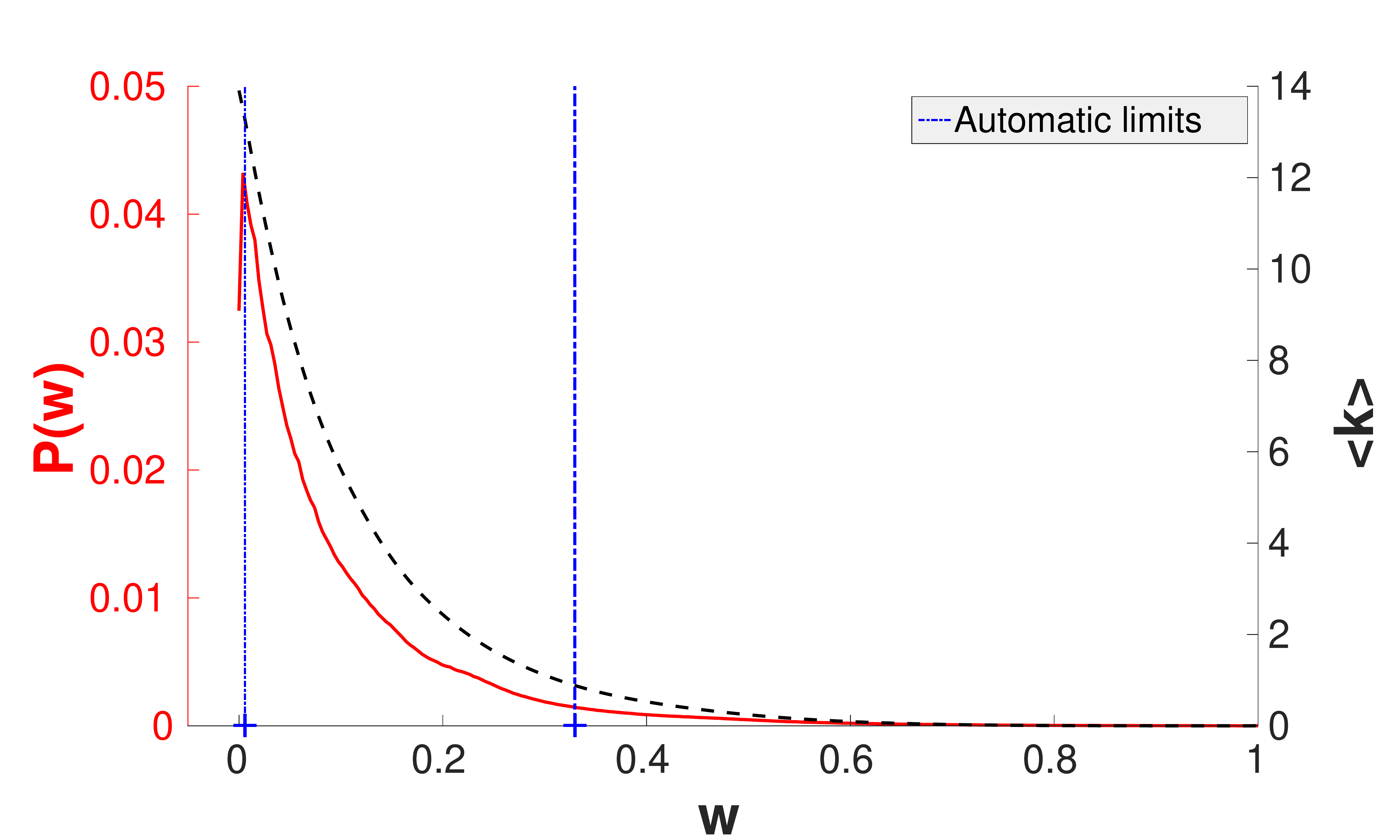}}\subfigure[Usptex dataset]{\includegraphics[width=0.33\linewidth]{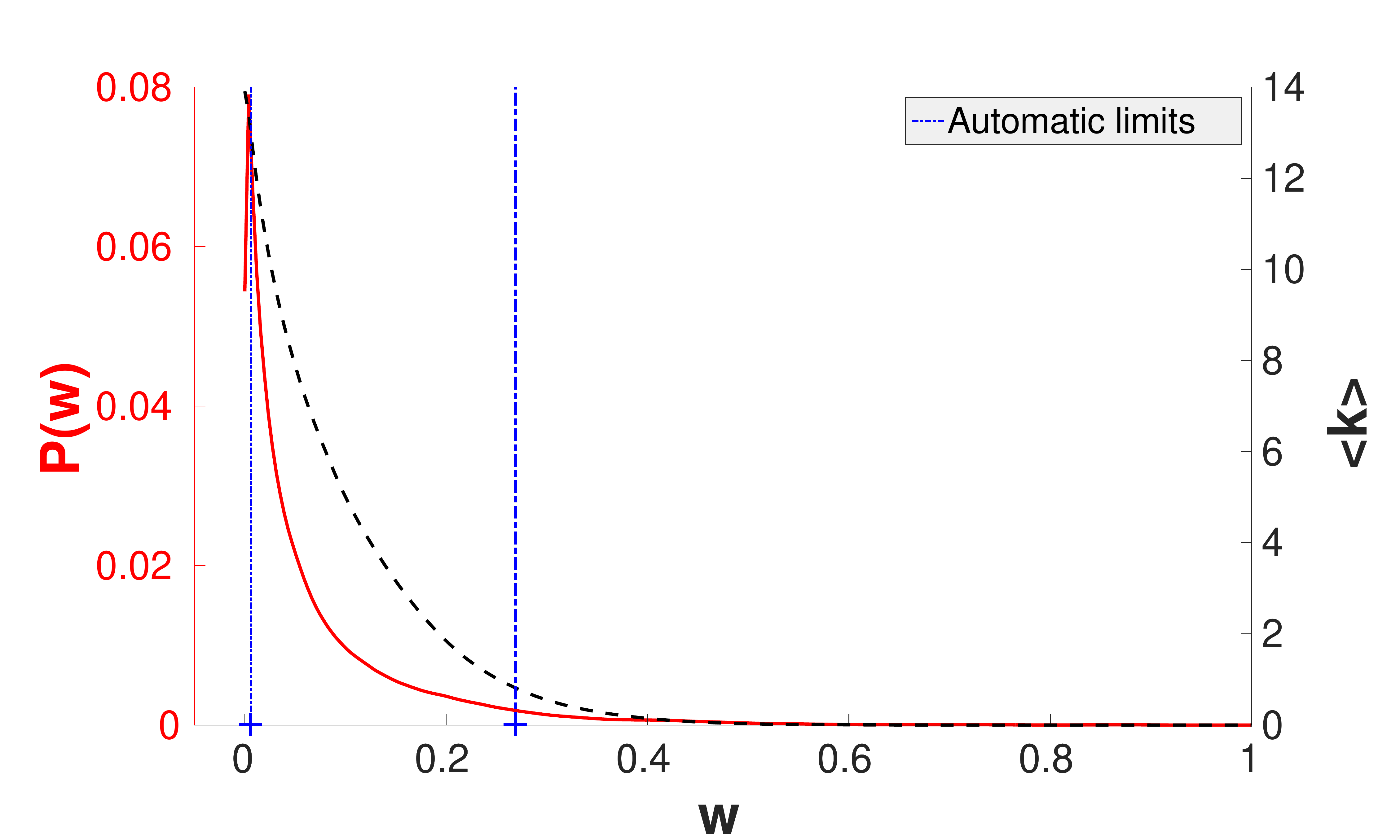}}
	\subfigure[Outex13 dataset]{\includegraphics[width=0.33\linewidth]{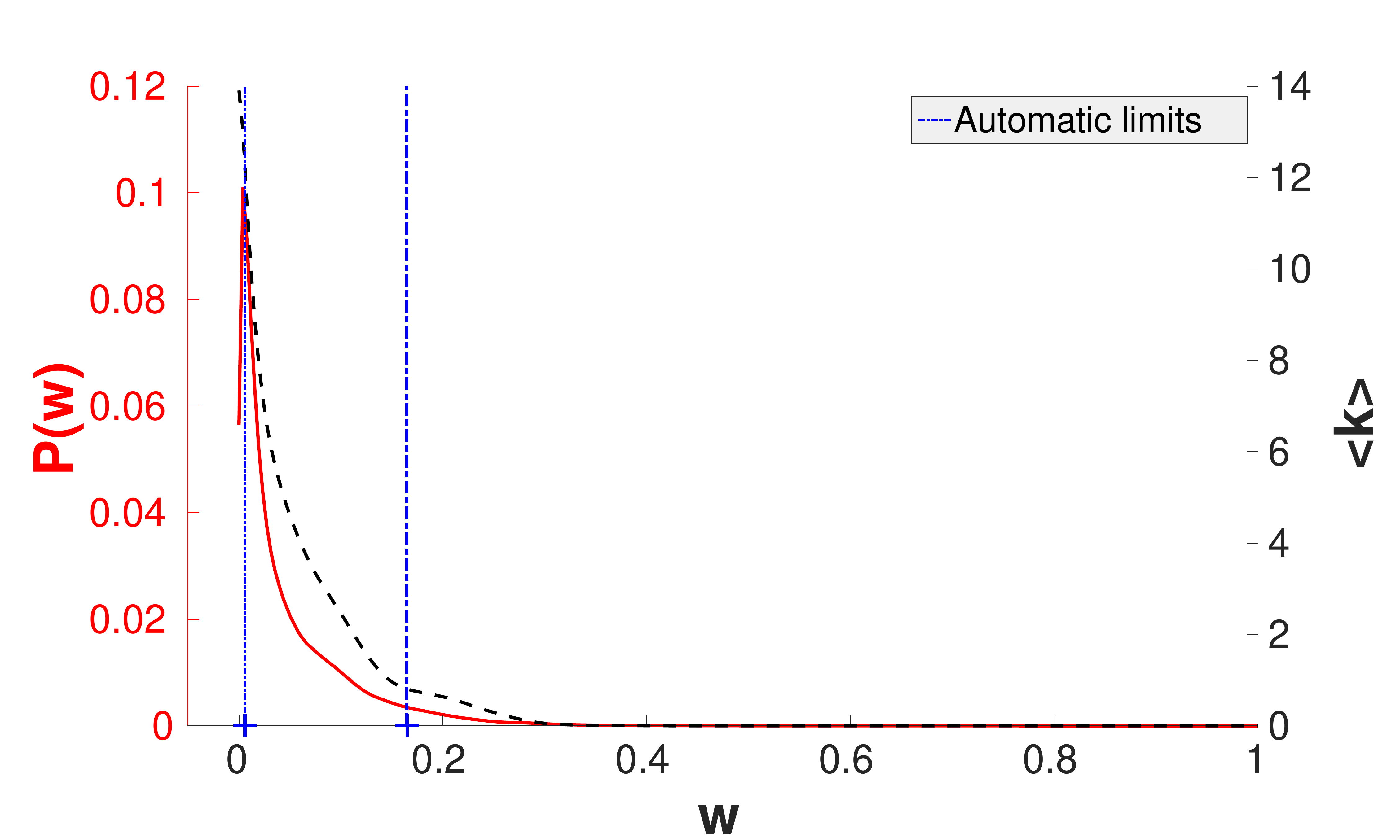}}
	\caption{\label{fig:edgesdistribution} Edge weight distribution $P(w)$ (red curve) of $N^{r_1}$ networks for 3 color-texture datasets. In black, the mean degree of all networks built using $w$ as threshold $N^{r_1, t=w}$. Blue lines represents the estimated threshold limits using the proposed automatic approach.} 
\end{figure}

We can draw important conclusions from the edge weight distribution. First: the distribution varies from one dataset to another. Empirical tests indicate that changing the radius does not imply in relevant changes on both curve behaviors. Second: as the mean, degree is the mean number of vertex connections, at the tails of the distribution we either have a fully connected network when $t=w \approx 0$ or too few connections ($\mu_k \leq 1$), when the threshold reaches values with low edge occurrence. Moreover, this happens in different moments from one dataset to another. Based on these findings we argue that using constant thresholds is not the best approach to model networks. In fact, to optimize results the user would need to evaluate the best threshold values for different kinds of image. 

To avoid the limitation of manually defining the threshold range, we propose an automatic approach to define $[t_1, t_m]$. We want to remove extreme cases when the network is almost/fully connected or when the network has 0 or too few connections, i.e. the tails of the distribution. First, we must define the lower inferior limit $t_1$ of the threshold interval. By analyzing the curves on Figure \ref{fig:edgesdistribution}, it is possible to notice that the peak of the function is usually around $[0, 0.1]$, region which results in networks with higher mean degree values. This happens because the lowest possible value of $w$ is $\frac{1}{L(r+1)}$, and most edges has weigh around $[\frac{1}{L(r+1)}, \frac{3}{L(r+1)}]$. These edges connect pixels of similar intensity values, which are predominant in most images. Therefore, as we are discarding low-weighted connections to highlight edges between different pixels, we can discard values smaller than the peak of the distribution. Thus the inferior limit of the threshold range is defined by

%The idea is to use the mean $\mu$ and standard deviation $\sigma$ of the edge weight distribution to access its limits. Consider a scenario with a set $Q$ of training images $I$ and a given radius $r$. First, we obtain all $N_I^r$ networks from the training images $I \in Q$ and extract its set of edges $E_I$. With the edge weight distribution, we then calculate its mean $\mu$:
%\begin{equation}
%    \mu = \frac{1}{|Q|*|E_I|}\ \sum_{I \in Q}\ \ \sum_{w(v_{i}, v_{j}) \in E_{I}}  (w(v_i, v_j))
%    \end{equation}

%and standard deviation $\sigma$:    

%\begin{equation}
%    \sigma = \sqrt{\frac{1}{|Q|*|E_I|}\ \sum_{I \in Q}\ \ \sum_{w(v_{i}, v_{j}) \in E_{I}} (w(v_i, v_j) - \mu)^2}
%\end{equation}
%where $|Q|$ is the number of training images and $|E_I|$ the number of edges of each $N_I^r$ network. The second step is to estimate the threshold limits ($[t_1, t_m]$). According to the 68–95–99.7 rule, $t$ values between $\mu - \alpha \sigma < t < \mu + \alpha \sigma$ covers a certain fraction of a normal distribution according to $\alpha$. For instance, $\alpha=1$ covers approximately $68\%$ of the distribution, while $\alpha=3$ covers $99.7\%$. This approach provides interesting results in defining the coverage limits for the threshold set, as we show in the following. We want to remove extreme cases when the network has 0 or too few connections, or when the network is fully/almost fully connected, i.e. the tails of the distribution. Therefore, we adopted $\alpha=2$ (covers $\approx 95\%$ according to the rule) to define the threshold set limits:

\begin{equation}    
t_1= \argmax_{w=0}^{1} (P(w))
\end{equation}
where the function $\argmax$ returns the value $w \in \rm I\!R$ that produce the maximum of function $P$.

As the left tail of $P$ relates to denser networks, the right tail produces sparse networks with a lower mean degree. Networks with mean degree $\leq 1$ does not provide relevant topological information for the characterization because the degree distribution tend to be sparse, with many disconnected vertices. To remove these networks, we want to find a moment $w=t$ necessary to satisfy $\mu_k \leq 1$. In other words, $w$ is the value of $t$ necessary to produce networks with mean degree $\leq 1$. Considering the function of the mean degree as $\gamma (w)= \mu_k$, according to the curve shown on Figure \ref{fig:edgesdistribution}, then

\begin{equation}\label{eq:tm2}
t_m = \argmax_{w=0}^{1} (\delta (\gamma(w), 1)) 
\end{equation}
where $\delta$ is the Kronecker delta (returns 1 when $\gamma(w) = 1$, and 0 otherwise). On this case, $\argmax$ returns the value of $w$ necessary to $\mu_k = 1$.

Computing the mean degree of networks from the training set for all $w=t$ can be costly, so we propose a practical way to find $[t_1, t_m]$ without it. First we analyze the integral $\int^{t_m} P(w) dw$, which is the fraction of samples covered until $t_m$. On Table \ref{tab:integrals} we show the computed values for $t_m$ and the results of the integral on 3 texture datasets. As the distributions are normalized, $\int P(w) dw = 1$. It is possible to notice that $t_m$ is different on each dataset, as it depends on the $P(w)$ distribution. By associating $t_m$ with the integral of $P$ it is possible to observe that independently of $t_m$, the transition $\mu_k \leq 1$ usually occur when around $90\%$ of the distribution is covered. Intuitively, as each dataset has a different $P(w)$ distribution, $t_1$ and $t_m$ limits will adapt to each case, but covering a similar fraction of the edges. Finally, to define a generic $t_m$ without knowing the mean degree of the training set, we can simply find the value that satisfies $ \int^{t_m} P(w) dw = 0.9$. The blue lines on Figure \ref{fig:edgesdistribution} shows the threshold range obtained by the proposed approach, where it adapts to different $P$ distributions. 

\begin{table}
	\centering
	\caption{Integral of the edge weight distribution (function $P(w)$) in the interval between the inferior threshold limit ($t_1$) and when the mean degree distribution transits to $\leq 1$ (moment $t_m$). Results indicate that the transition usually happen at different points of the distribution ($t_m$), but when the integral is around $0.9$.}
	\begin{tabular}{|c|ccc|}
		\hline
		& Vistex& Usptex & Outex13 \\ 
		$t_m$ & 0.33 & 0.27 & 0.16\\
		
		$\int^{t_m} P(w) dw$& 0.90 & 0.89 & 0.89 \\    
		\hline            
	\end{tabular}      
	\label{tab:integrals}
\end{table}

%To show that the computed limits fit similarly to the coverage stated by the 68–95–99.7 rule, we compute the integral of $P(w)$ on the interval $[t_1, t_m]$, results to each dataset are shown in Table.

%The result for the Vistex dataset is the most similar according to the rule, covering $0.92$ ($92\%$) of the edge weight distribution. For the Usptex and Outex13 datasets, the proposed automatic thresholds cover $0.89$.

Here we used all the dataset images to show the edge weight distribution differences and compute the threshold limits. However, it is important to notice that in a realistic classification scenario, only a fraction of the images is available for training. Our experiments indicate that the number of images used to estimate the threshold limits has a minimal influence on the result. Using at least 1 image of each class is enough to have robust values. On the classification experiments conducted on Section \ref{sec:experiments}, we use the training set to estimate the threshold limits.

Using the threshold limits automatic estimated, we build the set $T$ by dividing the interval $[t_1, t_m]$ into $m$ equidistant values. Thus the user only needs to inform the number of thresholds desired ($m$). We suggest the use of at least $m=4$, up to $m=10$ for optimized results. Combining the sets $R$ and $T$ to model a set of CNs $N^{r, t}$ allows to analyze its dynamics, which imply in a wider possibility of characterization. We address this process in the next section.

\subsection{Complex Network Characterization}\label{sec:characterization} 

The pattern of connections of the CN $N^{r, t}$ contains important information about the color-texture it represents. Various approaches can be employed to quantify the topology of $N$, and we can also analyze specific edges separately. We propose to derive $N=\{V_N, E_N\}$ into 2 subnets $W$ and $B$, where $E_N = E_W \cup E_B$ and $V_N = V_W = V_B$. For that, for each edge $w$, if it connects two vertices in the same layer/color-channel then a new network $W^{r, t}=\{V, E_W\}$ is obtained from $N^{r, t}$

\begin{equation}\label{eq:W}
E_W = \forall w(v_i, v_j) \in E_N \left\{
\begin{array}{l}
w(v_i, v_j) $, \ if $p(v_i, z) = p(v_j, z)
\\
\ \ \ \ \ \emptyset $,  \ \ \ \ \ \ \ \ \ otherwise$
\end{array}
\right
.
\end{equation}
where $p(v_i, z)$ returns the color-channel of the pixel that $v_i$ represents. Similarly, edges connecting vertices in different layers are assigned to a new network $B^{r, t}=\{V, E_B\}$, that is also a subnet of $N^{r,t}$:

\begin{equation}\label{eq:B}
E_B = \forall w(v_i, v_j) \in E_N \left\{
\begin{array}{l}
w(v_i, v_j) $, \ if $p(v_i, z) \neq p(v_j, z)
\\
\ \ \ \ \ \emptyset $,  \ \ \ \ \ \ \ \ \ otherwise$
\end{array}
\right
.
\end{equation}

Each subnet $W$ and $B$ of $N$ are illustrated on Figure \ref{fig:in-out}. The network $W$ enhance patterns of intensity variation within a single color-channel, allowing to analyze the spatial behavior on each color individually. On the other hand, the network $B$ represents the relation of spectral variance between different color channels. This is somehow similar to the color opponent theory of the human visual system \cite{foster1895text}, which states that specific cells in the brain process variations between pairs of colors, rather than each color individually.

\begin{figure}
	\centering
	\subfigure[Original CN ($N^{r, t}$)]{\includegraphics[angle=-90, width=0.37\linewidth]{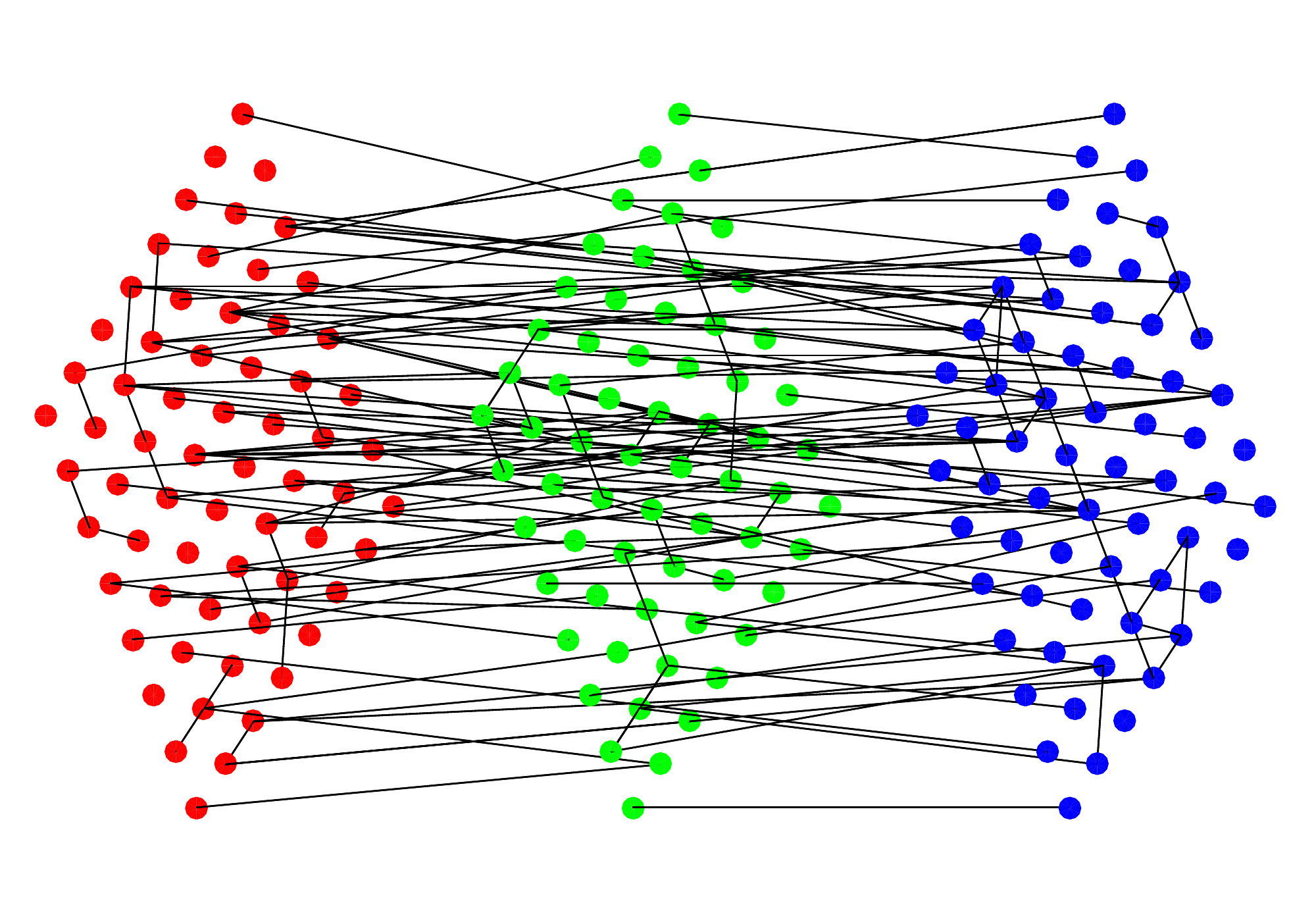}}\\
	\subfigure[Within-channel connections ($W^{r, t}$)]{\includegraphics[angle=-90, width=0.35\linewidth]{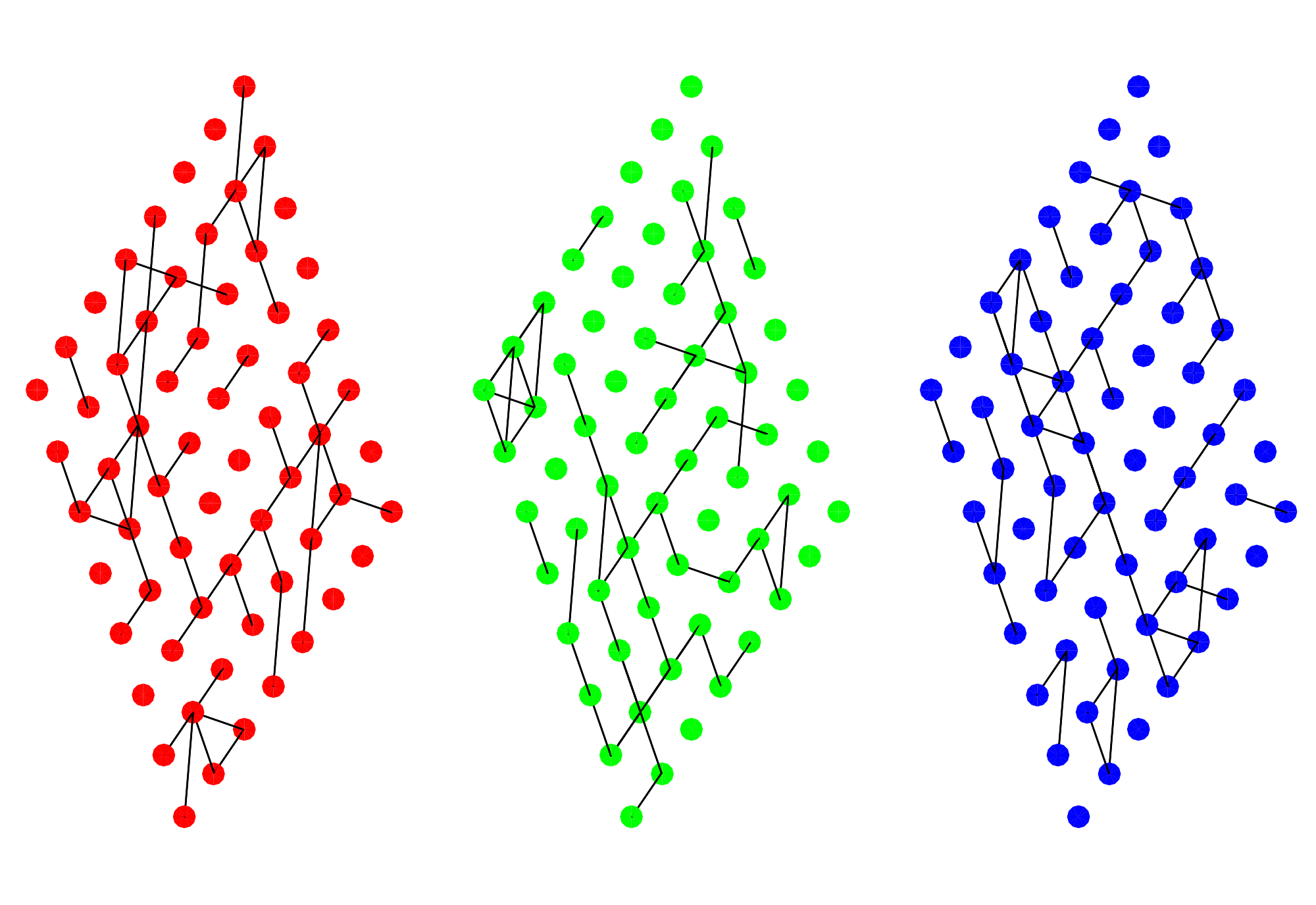}}
	\subfigure[Between-channel connections ($B^{r, t}$)]{\includegraphics[angle=-90, width=0.35\linewidth]{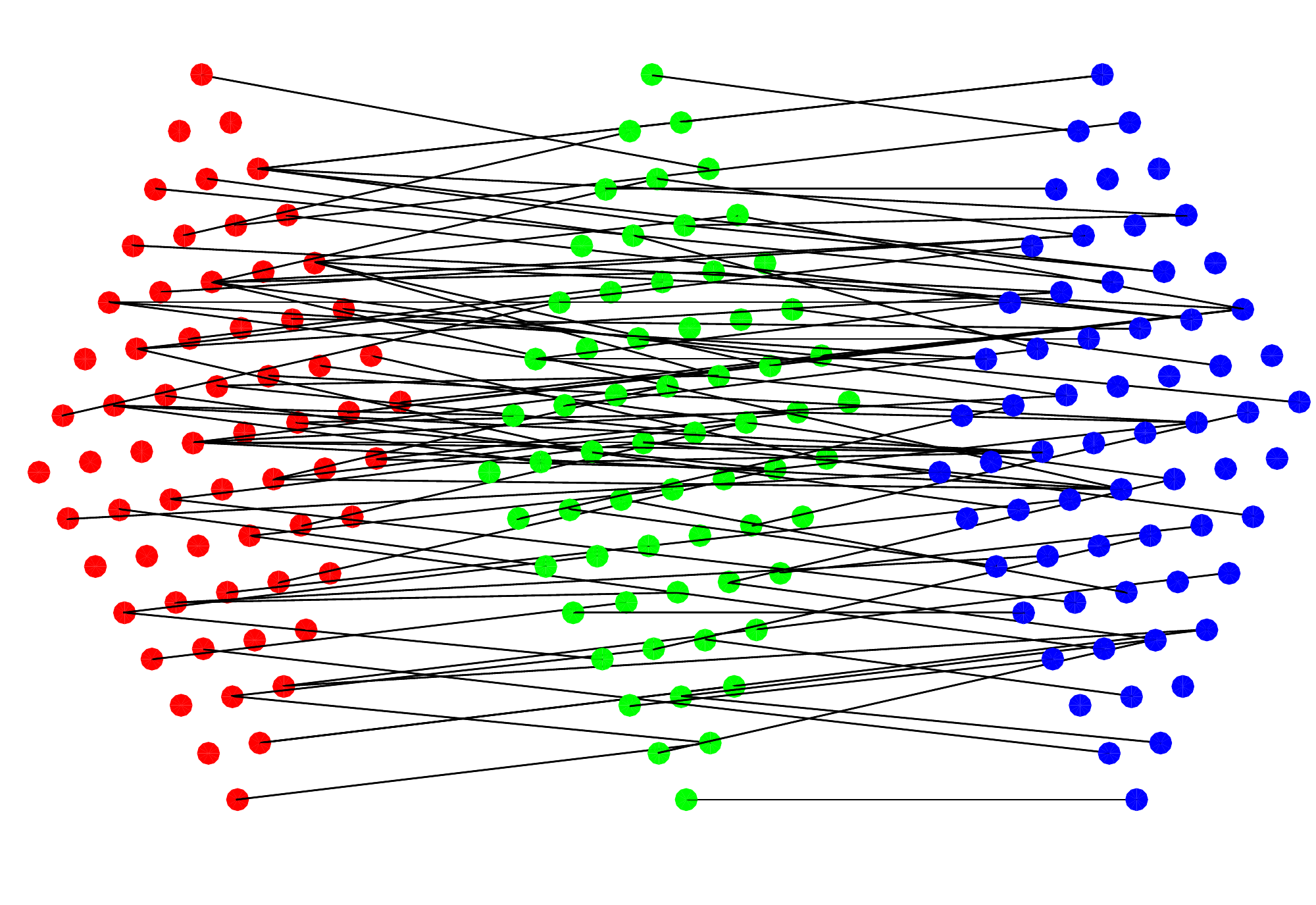}}
	\caption{\label{fig:in-out} Proposed technique to derive the structure of the original CN $N$ (a) in two subnets of connections within $W$ (b) and between $B$ (c) color-channels.}
\end{figure}

%The analysis of $W$ alone is similar to the analysis of common single-channel methods, which characterize each channel individually. In fact, the structure of the network $W$ is similar to a network obtained by using previous CN methods for gray-level texture analysis \cite{chalumeau2006,chalumeau2009,backes,scabini2015texture}. The only difference is the edge weight and the thresholding, as previous works remove high weighted connections (we remove low weighted connections, see Equation \ref{eq:cuting}).

%On Section \ref{sec:experiments} we show that the performance of using only the network $W$ is inferior to using $B$ or $N$. However, combining $W$ either with $B$ or $N$ (or both) improves results in most cases.

Topological features can be obtained from each network $N$, $W$ or $B$. One can find several kinds of measures that can be extracted from a CN (See \cite{structureanddynamics} for a review of measures). The major issue with most of them is its computational cost. Common measures such as closeness and betweenness have complexity $O(n*|E|)$, which is impractical in our case because $n=|V|=x*y*z$ and usually $|E| \gg |V|$, so the complexity is higher than $O(n^2)$. To improve the efficiency of the method (see Section \ref{sec:complexity} for detailed information), we focus on metrics with a lower computational order. We chose the degree $k$ (Equation \ref{eq:degree}), which can be computed while we build the network with cost $O(n*r_{neigh})$, where $r_{neigh}$ is the neighborhood size we visit while connecting vertices. Another measure chosen is the clustering coefficient $c$ of a vertex (Equation \ref{eq:clustering}), that can be computed with $O(n*\mu_k)$, where $\mu_k$ is the mean degree of the network, which usually is $\mu_k \ll n$ using the radii we considered on this work. It is important to notice that we empirically tested the characterization using only the degree or the clustering individually, but their combination proved to be more effective.

The degree and clustering of networks $N$, $W$ and $B$ capture interesting texture properties, as we show in Figure \ref{fig:visualA}. On this image, the CN $N^{r,t}$ is modeled using $r=4$ and $t=0.08$. The images are built by converting the 3 CN measures related to each pixel into an RGB intensity vector, and values are normalized to $[0,255]$. Regions with dark colors mean that the CN measure is low on the 3 layers of the network, while white regions mean high measures. Colored regions depend on intermediate values of each network layer. It is possible to observe that the network $W$ works similarly to a border detector. This is somehow similar to standard integrative methods of texture analysis, which capt patterns of intensity variation in a single channel. Moreover, the degree and clustering seem to highlight the same overall border pattern, but with small differences in some regions of the image. The degree highlight soft and small borders, while the clustering seems to have a higher tolerance to variations, highlighting fewer borders. This happens because the calculation of the clustering coefficient considers the connections between the neighbors of the vertex, while the degree only considers the connections of the vertex. In other words, to be higher the clustering needs a higher interconnectivity among a group of neighbor vertices, not just the vertex in question. We can also notice that the network $W$ enhance a wider range of color variations on the left texture sample, which happens frequently in all color channels of this image. On the other hand, on the leaf sample, the gradient is higher as whiter/pink regions transits to dark regions, and these variations are effectively highlighted by the CN measures.

\begin{figure}
	\centering
	\subfigure[Input textures]{\includegraphics[width=0.1\linewidth]{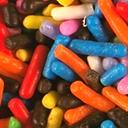} \ \ \ \ \ \ \ \    \includegraphics[width=0.1\linewidth]{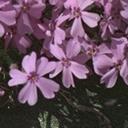}}\\
	\subfigure[degree $k$ of $N^{r,t}$]{\includegraphics[width=0.1\linewidth]{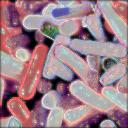} \ \ \ \includegraphics[width=0.1\linewidth]{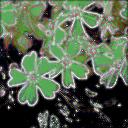}} \ \ \ \ \ \ \ \ \ \ \ \ \ \ \ \subfigure[clustering $c$ of $N^{r,t}$]{\includegraphics[width=0.1\linewidth]{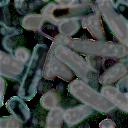} \ \ \ \includegraphics[width=0.1\linewidth] {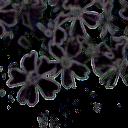}}      
	\\
	\subfigure[degree $k$ of $W^{r,t}$]{\includegraphics[width=0.1\linewidth]{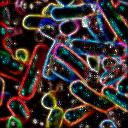} \ \ \ \includegraphics[width=0.1\linewidth] {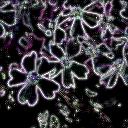}} \ \ \ \ \ \ \ \ \ \ \ \ \ \ \ \subfigure[clustering $c$ of $W^{r,t}$]{\includegraphics[width=0.1\linewidth]{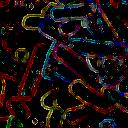} \ \ \ \includegraphics[width=0.1\linewidth] {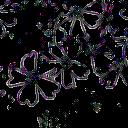}}\\
	\subfigure[degree $k$ of $B^{r,t}$]{\includegraphics[width=0.1\linewidth]{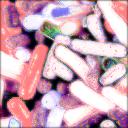} \ \ \ \includegraphics[width=0.1\linewidth]{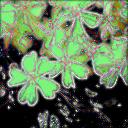}} \ \ \ \ \ \ \ \ \ \ \ \ \ \ \ \subfigure[clustering $c$ of $B^{r,t}$]{\includegraphics[width=0.1\linewidth]{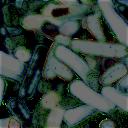} \ \ \ \includegraphics[width=0.1\linewidth]{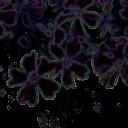}}\\
	
	\caption{\label{fig:visualA} Qualitative analysis of the proposed method where the CN topological measures (degree and clustering) are converted into RGB images. The networks are modeled using the input image (a) with $r=4$ and $t=0.08$. Measures are converted into an RGB intensity vector by concatenating the 3 values of vertices related to each pixel (1 value per layer/color-channel). Values are normalized into $[0,255]$, where low values mean low CN measures.}
\end{figure}

The patterns highlighted by the network $B$ are different from the network $W$. The degree and clustering performed differently on the left texture sample. Considering the degree, it seems that the variations between channels are predominant on specific channels, i.e. one color-channel remains constant while the others vary. For instance, on the left texture sample, the connectivity of the vertices is either a bitter higher on the red or on all color-channels (white regions), with some exceptions at small blue and green regions. This also happens on the plant texture sample, with predominant connectivity on the green color-channel. This indicates that the CN here capt regions with higher variations with patterns more related to the overall color, rather than local changes (sharp borders), as capt by the network $W$. On the other hand, the clustering coefficient response is not predominant in a single color-channel. Thus, it either highlights regions where the coefficient is high or low in all channels. On the left texture sample, it worked similarly to the degree, but with small changes in the predominant response from the red channel to the green. On the plant image, it works similarly to the clustering obtained with the network $W$, highlighting patterns of border. The network $N$ seems a combination of $W$ and $B$ with some particularities. All these networks along with each topological measure capt important color-texture information that allows an effective characterization of the images.

%In other words, a relevant intensity change must happen simultaneously on all color channels for vertices to have high clustering.

Finally, to characterize the CNs we summarize the degree and clustering distributions using 4 traditional statistics from the literature, the mean, standard deviation, energy and entropy, equations are given on Table \ref{tab:statistics}. It is important to notice that the clustering coefficient cannot be computed when considering within-channel connections (network $W$) and $r=1$. This happens because the Euclidean distance 1 is not enough for triangles to form in this case. Therefore, we discard the clustering features for this case as they are always 0. 

\begin{table}
	\centering 
	\caption{Equations of the statistical measures used to summarize the degree and clustering distributions. Each measure can be computed for one of the 3 networks $N^{r,t}$, $W^{r,t}$ or $B^{r,t}$. Measures are extracted for both the degree $k$ ($\mu_k$, $\sigma_k$, $e_k$ and $\epsilon_k$) and the clustering ($\mu_c$, $\sigma_c$, $e_c$ and $\epsilon_c$) distributions. $P$ is a probability density function, so $P(k)$ is the probability of degree $k$ to occur on the network, and $k(i)$ is the degree of vertex $i$ (analogous for $P(c)$ and $c(i)$).}
	
	\begin{tabular}{|c|c|}
		\hline
		Mean &  Standard deviation \\         
		$\mu_k = \frac{1}{|V|} \sum\limits_{i=1}^{|V|} k(i)$ & $ \sigma_k = \sqrt{\sum\limits_{i=1}^{|V|} ( k(i) - \mu_k)^2} $\\
		\hline
		Energy & Entropy\\
		$e_k = \sum\limits_{k} P(k)^2$ & $\epsilon_k = - \sum\limits_{k} P(k)*\log (P(k))$\\
		\hline
		
	\end{tabular}      
	\label{tab:statistics}
\end{table}

A feature vector is built to characterize color-texture by concatenating several parameters and measures from the CNs. First, given a network ($N^{r,t}$, $W^{r,t}$ or $B^{r,t}$), the 4 statistical measures from the degree and clustering distributions are combined into a vector $\varpi$:

\begin{equation}\label{eq:statistics}
\varpi(N^{r,t}) = [\mu_k, \sigma_k, e_k, \epsilon_k, \mu_c, \sigma_c , e_c , \epsilon_c]
\end{equation}

According to the dynamic analysis previously described (section \ref{sec:automatict}), we concatenate features of combinations of the sets $R$ and $T$, which yields a total of $i*m$ combinations. Given a network $N$, $W$ or $B$, a feature vector of size $|\varphi| = 2*4*i*m$ is then obtained, where 2 is the number of CN measures ($k$ and $c$) and 4 the number of statistics used from each measure:

\begin{equation}
\varphi_N = [\varpi(N^{r_1,t_1}), ..., \varpi(N^{r_i,t_m})]
\end{equation}

\begin{equation}
\varphi_W = [\varpi(W^{r_1,t_1}), ..., \varpi(W^{r_i,t_m})]
\end{equation}

\begin{equation}
\varphi_B = [\varpi(B^{r_1,t_1}), ..., \varpi(B^{r_i,t_m})]
\end{equation}

% \begin{equation}
% \varphi = [\varphi_N, \varphi_W, \varphi_B]
% \end{equation}

The final Multilayer CN descriptors can be either each one of the feature vectors for $N$, $W$ or $B$ or their combination:

\begin{equation}
\varphi = [\varphi_N, \varphi_W, \varphi_B]
\end{equation}

We evaluate the performance of each individual $\varphi$ and some different combinations on Section \ref{panalysis}. All steps to model and characterize a color-texture with the proposed method are illustrated in Figure \ref{fig:method}.

\begin{figure}
	\centering
	\includegraphics[width=0.4\linewidth]{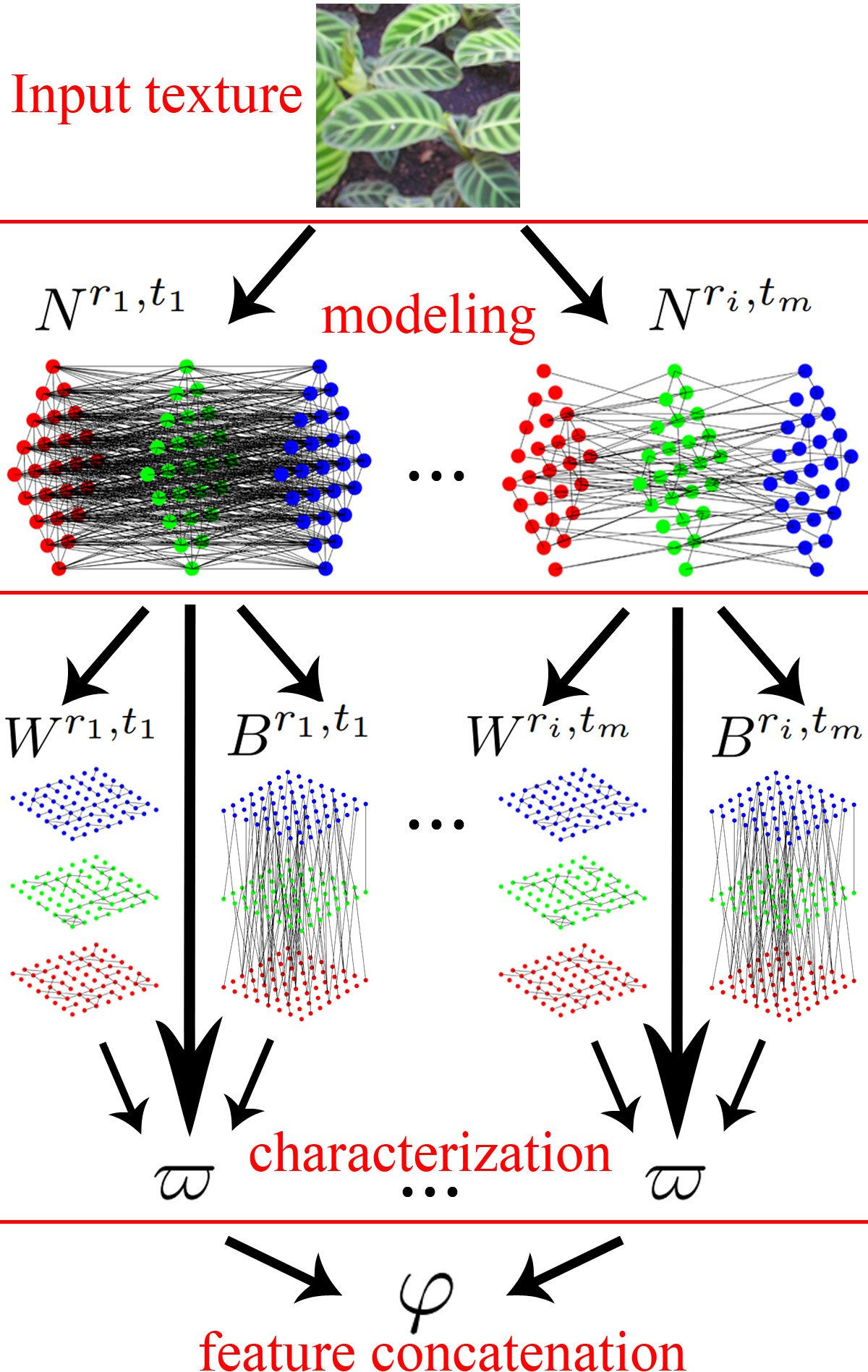}
	\caption{\label{fig:method} Step-by-step of the proposed method. The modeling step considers the network dynamic evolution, thus combining each parameter $r$ and $t$ results in a set of networks $N$. The characterization consists on first obtain the subnets $W$ and $B$ and then quantify the structure of all networks ($N$, $W$ and $B$) with degree ($k$) and clustering ($c$) statistics $\varpi$ (see Table \ref{tab:statistics} and Equation \ref{eq:statistics}). The final feature vector $\varphi$ is the concatenation of statistics from all networks.}
\end{figure}

\subsection{Computational Complexity}\label{sec:complexity}

Here we discuss the total computational complexity to obtain the Multilayer CN descriptors, from the modeling of the networks to their characterization. On the modeling step, it is necessary to visit each pixel, in each color-channel, and check its neighborhood to create the network connections. Thus, considering an image $I$ with size $x*y$ and $z$ color-channels, the modeling of a network $N^{r,t}=\{V, E\}$ takes $O(n*r_{neigh})$ where $r_{neigh}$ is the neighborhood covered by an radius $r$ and $n=x*y*z$. The networks $W$ and $B$ are obtained during the modeling process by just checking the corresponding color-channel of the pixels when defining a connection with Equation \ref{eq:connectionWeight}. Now considering the dynamic network analysis, the modeling process is repeated concatenating all values of the sets $R=\{r_1,..., r_i\}$ and $T=\{t_1,..., t_m\}$. The whole modeling process then has complexity $O((n*r_{neigh})*i*t)$. We can remove lower order terms that are much smaller than $n$. Usually, $i\leq6$ and $m\leq14$ (highest values we analyze in this work), so we can discard these terms from the equation. The neighborhood covered by the radius varies depending on $r$. For instance, $r=1$ covers $r_{neigh}=14$, and $r=3$ covers $r_{neigh}=38$. Usually, this number is also much smaller than $n$ (e.g. an image from the Usptex dataset, with size $200*200$ and $z=3$, has $n=120000$). Obviously, if one wants to use a radius that covers larger parts of the image (i.e. $r_{neigh} \approx n$), the complexity then tends to $O(n^2)$. Therefore, we do not suggest that because empirical tests indicate that the performance stabilizes around $r=[4,6]$.

%By removing these lower order terms from the equation, we have a modeling complexity of $O(n)$.

The complexity of the network characterization depends on the cost to compute the topological measures. The degree $k$ distribution of the network can be obtained while we visit vertices to create connections, in the modeling step. The clustering coefficient must be computed after the network modeling and has complexity $O(n*\mu_k)$ where $\mu_k$ is the mean degree of the network. The mean degree can never be higher than the size of the neighborhood visited to create connections, so $\mu_k \leq r_{neigh}$. Finally, by summing the modeling and the characterization complexity the asymptotic limit of the proposed method is:
%Thus, we can discard this term from the equation, resulting in $O(n)$. 

\begin{equation}
O(2n r_{neigh}) = O(2n \mu_k)
\end{equation}

%where $n$ is the size of the network, which is the number of pixels of the image times the number of color channels ($n= |V| = x*y*z$).

\section{Experiments and Discussion}\label{sec:experiments}

This section presents several experiments to analyze the performance of the proposed method and its relevance in comparison to methods from the literature. The experiments consist of classification tasks in 4 color-texture datasets of RGB images. Although evidence raised by some works \cite{paschos2001perceptually, choice2011, cernadas2017} points that classification results can be improved by using different color spaces such as HSV and Lab, we decided to approach the traditional RGB space for simplicity purposes. Our focus on this work is on the modeling and characterization of color-texture, regardless of the chosen color space. As a future work, further investigation on the pertinent choice of color space and their influence on the method should be considered. 

\subsection{Datasets}\label{datasets}

The following color-texture datasets are used on this work:

\begin{itemize}
	\item Vistex: The Vision texture dataset \cite{vistex} has 54 natural color images of size 512x512. Each image of the original 54 classes is split into 16 non-overlapping sub-images of 128x128 pixels, resulting in 864 images. %Some samples of the first 20 Vistex classes are shown on Figure \ref{fig:datasets} (a). 
	
	\item Usptex: The Usptex dataset \cite{backes2012}, from the University of São Paulo, is a larger dataset with 191 classes of natural textures, that are commonly found daily. The original 191 images have 512x384 pixels and are split into 12 non-overlapping 128x128 sub-images, totalizing 2292 images. %Samples of 20 classes from Usptex are shown on Figure \ref{fig:datasets} (b).
	
	\item Outex13: The Outex framework \cite{ojala2002outex} was proposed for the empirical evaluation of texture analysis methods. At the Outex site, the Outex13 dataset is the test suite Outex$\_$TC$\_$00013, which address color as a discriminative texture property. It contains 1360 color images of 68 texture classes, with 20 samples each. %On Figure \ref{fig:datasets} (c) we show 20 classes from the Outex13 dataset.
	
	\item CURet: This is a color dataset of material images from Columbia-Utrecht, and contains 61 texture classes with 92 samples each \cite{Varma2005}. Images have considerable within-class variations such as rotation, illumination and view angle. %A preview of 20 CURet classes is shown on Figure \ref{fig:datasets} (d).
	
	\item MBT: The Multi Band Texture dataset \cite{abdelmounaime2013newbrodatz} provides colored images formed from the combined effects of intraband and interband spatial variations, a behavior that appears on images with high spatial resolution. Images with this property usually derive from areas such as astronomy and remote sensing. To compose the dataset, each one of the 154 original MBT classes, with sizes 640x640, was split into 16 non-overlapping samples of size 160x160, resulting in 2464 images in total. %Some samples are shown on \ref{fig:datasets} (e).
\end{itemize}

% \begin{figure}
%        \centering
%       \subfigure[Vistex]{\includegraphics[width=0.3\linewidth]{vistexMosaic.jpg}} \ \ \ \ \ \ \ \ \   \subfigure[Usptex]{\includegraphics[width=0.3\linewidth]{usptexMosaic.jpg}}
%       \subfigure[Outex13]{\includegraphics[width=0.3\linewidth]{outex13Mosaic.jpg}} \ \ \ \ \ \ \ \ \   \subfigure[CURet]{\includegraphics[width=0.3\linewidth]{curetMosaic.jpg}}\\
%       \subfigure[MBT]{\includegraphics[width=0.33\linewidth]{mbtMosaic.jpg}} 
%         \caption{\label{fig:datasets} Samples of 20 classes from each dataset used in the experiments of this work.}
% \end{figure}

\subsection{Classification Setup}

For classification, we use the Linear Discriminant Analysis (LDA) classifier \cite{LDA}. The LDA is a supervised method that finds a projection of the features where the between-class variance is larger compared to the within-class variance. %We empirically found that LDA seems to be the better choice for our method. In many cases, the performance of the classifier depends on the features of the method in question, as different classification paradigms fit better on specific data. Therefore, some works employ other classifiers such as k-nearest-neighbors (KNN) and variations of the support-vector-machine (SVM). 
In each experiment, the dataset is split into a leave-one-out scheme, where 1 sample is used for test and the rest is used for training. This process is repeated until all samples are used for test. Results are measured by the classification accuracy, which means the rate of images correctly classified. 

\subsection{Proposed Method Analysis}\label{panalysis}

We first analyze the influence of each modeling parameter on the performance of the proposed method. Figure \ref{fig:experiment-nt} shows the classification accuracy on 3 image datasets by using several parameter combinations. We analyze the performance of each radius $R=\{1,2,3,4,5,6\}$, individually or combined, by varying the number of thresholds $m$ from 2 to 14 using networks $N$ ($\varphi_N$). The first conclusion we can draw from these results is that the combination of 2 or more radius is more effective than using any individual radius. In most cases, the best results are achieved by combining 4 or more radius. As for the number of thresholds, it is possible to notice that $m=2$ provides the worst results, in all cases. The performance stabilizes after $m=4$ on the Vistex dataset and after $m=8$ on Usptex and Outex13. According to these results, we believe that $m=10$ is enough to have a robust performance in different scenarios, so we use that in the following experiments. 

\begin{figure}
	\centering
	\subfigure[Vistex]{\includegraphics[width=0.31\linewidth]{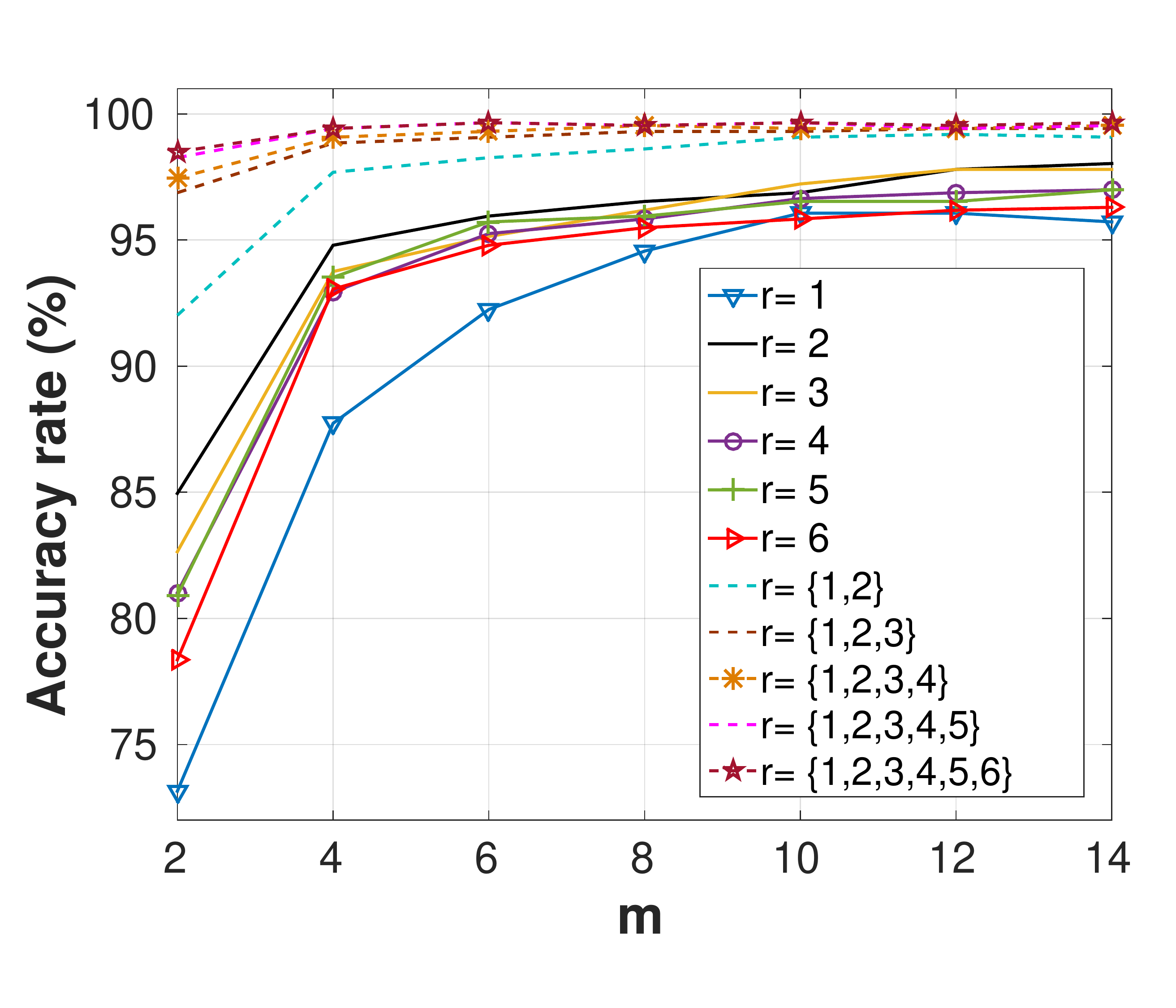}} \subfigure[Usptex]{\includegraphics[width=0.3\linewidth]{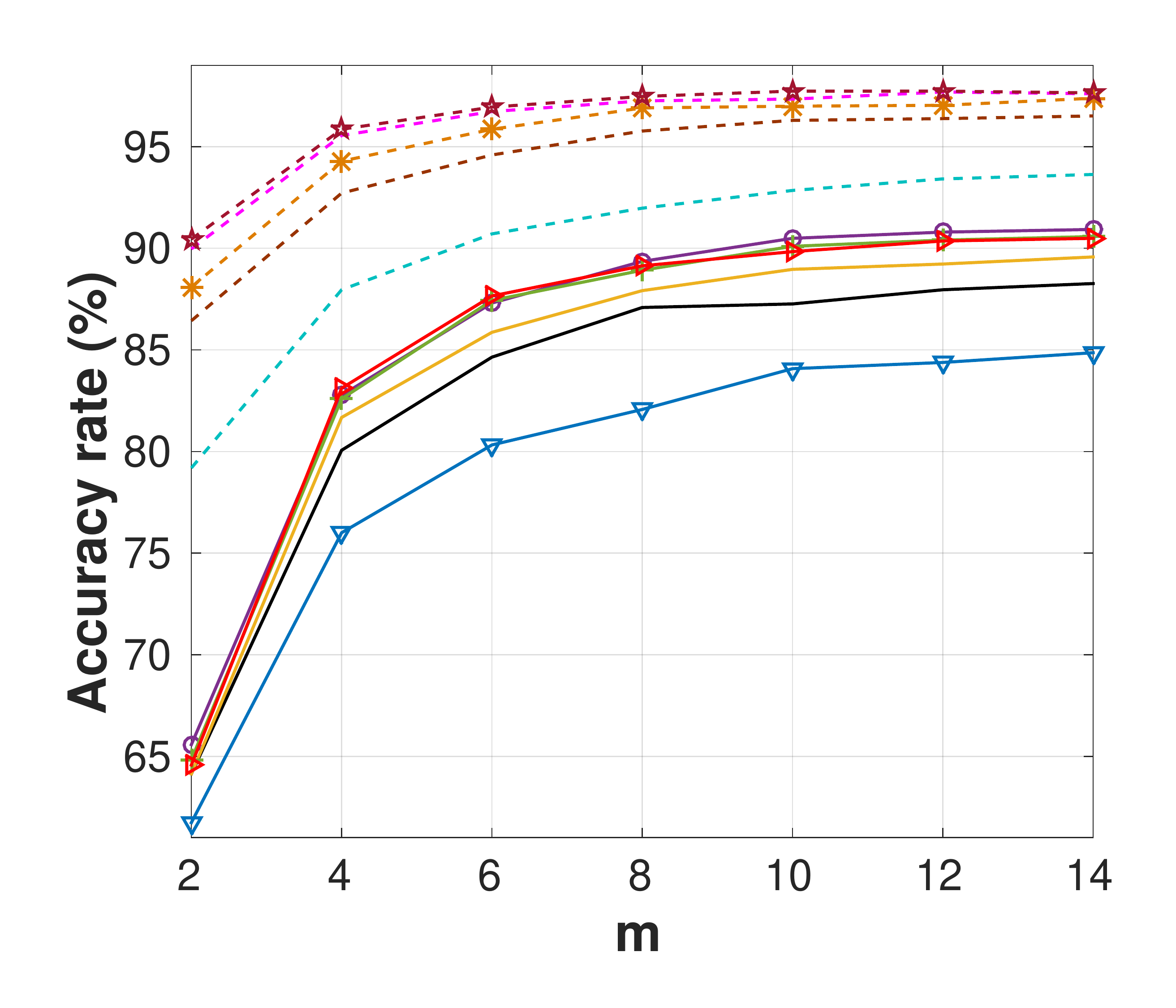}}
	\subfigure[Outex13]{\includegraphics[width=0.3\linewidth]{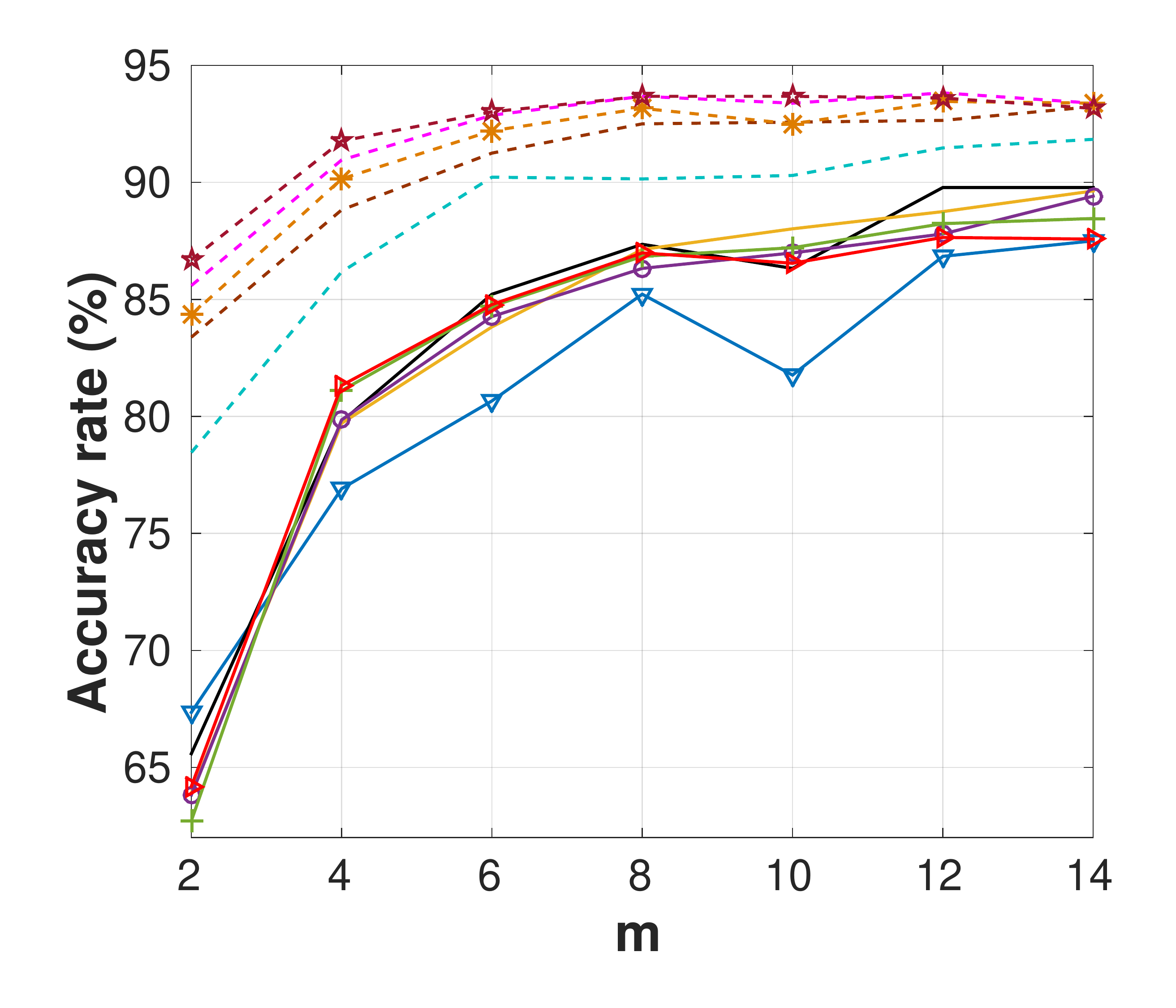}}
	\caption{\label{fig:experiment-nt} Classification accuracy rates with various combination of parameters on 3 datasets. Each curve represents the performance using a different radius $r$ (or their combinations) by varying the number of thresholds $m$ from 2 to 14.}
\end{figure}

The following experiment evaluates the performance of features from the different subnets $W$ ($\varphi_W$) and $B$ ($\varphi_B$) of $N$. Classification is performed using each network, individually or combined, with $m=10$ and a combination of 4 or more radius, until 6. Results for all the datasets are given on Table \ref{tab:INOUT}. For Vistex, the entire within-between channel analysis (network $N$) achieves the highest accuracy rate of $99.9$ using 5 or 6 radii. The lowest results are achieved using only the network $W$, however, the combination of $W$ with $B$ provides best results compared to $B$ alone. The combination of the 3 networks achieves an accuracy rate of $99.8$ using 3 radii, but the performance drops if we increase it.

\begin{table}
	\caption{Classification results of the proposed method on 5 color-texture datasets by combining topological features from different networks ($N$, $W$ and $B$). It was used $m=10$ and different numbers of radius, from 4 to 6. }\label{tab:INOUT}
	\centering
	\resizebox{\linewidth}{!}{
		\begin{tabular}{cc|c|c|c|c|c|c|}
			
			&&$\varphi =[\varphi_W]$ & $ \varphi = [\varphi_B]$ & $\varphi =[\varphi_W, \varphi_B]$&$\varphi =[\varphi_N]$ & $\varphi =[\varphi_N, \varphi_W, \varphi_B]$\\
			\hline
			%$R=\{1\}$           &88.7&    93.8&    98.4&    95.0&    99.0\\
			%\multirow{5}{*}{\rotatebox[origin=c]{90}{\textbf{Vistex}}}&$R=\{1,2\}$         &97.0&    98.5&    99.7&    99.1&    99.7\\
			%&$R=\{1,2,3\}$       &98.6&    99.4&    99.8&    99.3&    99.8\\   
			\multirow{3}{*}{\rotatebox[origin=c]{90}{\textbf{Vistex}}}&$R=\{1,2,3,4\}$     &99.0&    99.8&    99.7&    99.7&    99.4\\   
			& $R=\{1,2,3,4,5\}$   &99.1&    99.5&    99.4&    \textbf{99.9}&    97.3\\
			&$R=\{1,2,3,4,5,6\}$ &99.0&    99.7&    98.8&    \textbf{99.9}&    97.8\\
			\hline
			%& $R=\{1,2\}$         &85.0&    92.8&    96.8&    93.6&    97.6\\    
			%&$R=\{1,2,3\}$       &91.4&    96.1&    98.2&    96.5&    98.6\\    
			\multirow{3}{*}{\rotatebox[origin=c]{90}{\textbf{Usptex}}}&$R=\{1,2,3,4\}$     &93.4&    97.2&    98.4&    97.4&    98.7\\    
			&$R=\{1,2,3,4,5\}$   &94.2&    97.2&    98.7&    97.6&    \textbf{99.0}\\
			&$R=\{1,2,3,4,5,6\}$ &94.4&    97.3&    98.7&    97.7&    98.9\\
			\hline
			% &$R=\{1,2\}$         &80.2&    90.4&    92.9&    91.3&    94.0\\    
			%&$R=\{1,2,3\}$       &85.4&    92.6&    93.6&    93.3&    94.8\\    
			\multirow{3}{*}{\rotatebox[origin=c]{90}{\textbf{Outex}}}&$R=\{1,2,3,4\}$     &86.9&    93.6&    94.7&    93.8&    94.3\\    
			&$R=\{1,2,3,4,5\}$   &87.9&    94.6&    95.2&    94.7&    94.5\\
			&$R=\{1,2,3,4,5,6\}$ &88.3&    94.7&    95.3&    \textbf{95.4}&    93.7\\
			\hline
			% &$R=\{1,2\}$        &79.0&    77.2&    87.2&    78.5&    89.4\\
			%&$R=\{1,2,3\}$      &86.4&    84.7&    92.7&    85.5&    94.2\\
			\multirow{3}{*}{\rotatebox[origin=c]{90}{\textbf{CURet}}}&$R=\{1,2,3,4\}$    &88.5&    88.7&    94.5&    88.1&    96.0\\
			&$R=\{1,2,3,4,5\}$  &90.4&    91.4&    95.5&    90.7&    96.8\\
			&$R=\{1,2,3,4,5,6\}$&91.0&    92.6&    96.1&    91.8&    \textbf{97.0}\\
			\hline
			% &$R=\{1,2\}$        &89.2&    73.0&    90.9&    80.9&    92.0\\
			%&$R=\{1,2,3\}$        &94.3&    81.2&    94.9&    87.0&    94.9\\
			\multirow{3}{*}{\rotatebox[origin=c]{90}{\textbf{MBT}}}&$R=\{1,2,3,4\}$        &96.0&    83.8&    96.0&   88.6&    96.0\\
			&$R=\{1,2,3,4,5\}$        &96.8&    85.1&    96.1&    89.5&    96.4\\
			&$R=\{1,2,3,4,5,6\}$        &\textbf{97.1}&    85.9&    96.7&    89.6&    96.1\\
			\hline
		\end{tabular}
	}
\end{table}

Unlike the Vistex dataset, on Usptex the combination of networks is better than using any network alone. The 3 networks combined provide the best results, using any number of radius, and the highest accuracy rate ($99.0$) is achieved using 5 radii. Similar to Vistex, on the Outex13 dataset the highest accuracy rate, $95.4$, is achieved using the network $N$ alone and 6 radii. On the other hand, the lowest results are achieved using only $W$. The combination of $W$ and $B$ achieve results close to the best, $95.2$ using 5 radii and $95.3$ using 6. Using less radius (2 or 3), the best results are achieved by combining all 3 networks. On the CURet dataset, the lowest results are achieved using each network alone, with small differences between them. The combination of networks seems to be the best approach also on this dataset. The highest results are achieved using the combination of all networks and 6 radii, with an accuracy rate of $97.0$. The results achieved on the MBT dataset varies from what should be expected considering the performance on the other datasets. Here, the use of $W$ alone proved to be more effective than any other approach, where the maximum classification accuracy achieved is $97.1$ with 6 radii. This highlight the property of this dataset, describing a different pattern of how color corroborates to form the texture of each class. The spatial analysis of each color-channel individually capt these patterns with a higher precision than analyzing the color-channel interaction. The use of $B$ or $N$ alone achieves considerably inferior results compared to $W$ (around 10\% less). When combining $B$ and/or $N$ with $W$, the performance drops by a small factor ($\approx$ 1\%), which corroborates that $W$ alone provides the most discriminant features.

The results analyzed here provides interesting insights about the property of the texture present on each dataset. First, observing the difference on using the spatial within-channel (network $W$) or the between-channel (network $B$) analysis, we can notice that on Vistex and CURet the difference of performance is minimal. This indicates a pattern of color-texture on Vistex and CURet in which within and between channel interaction behaves similarly, complementing each other. On Usptex, the performance using the between-channel analysis is a bit higher than using the within-channel analysis. The overall performance of $N$ and $B$ is similar, and the highest results appear when combining all approaches, which suggests that each kind of spatial analysis provides different relevant color-texture information.

The differences between $W$ and $B$ on Outex13 and MBT are more evident. The between-channel analysis is far more effective on Outex13, while the opposite happens on MBT. Moreover, using $N$ or combining $W$ with $B$ improved the performance on Outex13 by a small factor, and again the opposite happens on MBT. This suggests that between-channel interaction has a higher contribution to discriminate the Outex13 classes, with a small contribution from within-channel spatial patterns. On the other hand, the key feature to discriminate classes of the MBT dataset seems to be spatial patterns on each individual color-channel. As a general conclusion, it seems that the color-texture phenomena of Outex13 and MBT are more complex than what is observed in the other datasets.

%The results obtained in all datasets corroborates that the use of $W$ alone is usually the worst case in terms of classification. This happens because the analysis of color-channels individually is less effective than analyzing their interaction ($B$) or the entire network ($N$). The combination of different networks or the use of only $N$ proved to be the best approaches to achieve higher classification results.

\subsection{Comparison with other methods}

On this section, we perform a comparison of classification results between different color-texture descriptors. As a baseline of performance, we considered a very simple approach to characterize color images with five statistical measures (mean, variance, entropy, third and fourth moments) computed for each color-channel. This method is known as Color statistics, or first-order statistics, thus it does not consider the spatial distribution of the pixels, neither the interaction between color-channel.

We then compare the best results achieved by the proposed Multilayer CN descriptors with other methods from the literature. For coherence in the comparisons, we considered only results that use descriptors computed from the RGB color-space and without color normalization. The following literature methods are considered in the comparisons: Pertinent color space and parametric spectral analysis, Qazi et al. (2011) \cite{choice2011}; Fractal measures of color-channels, Backes et al. (2012) \cite{backes2012}; Shortest paths on graphs, Sá Junior et al. (2014) \cite{sajunior2014} and; Fractal measure of the mutual interference of color-channels, Casanova et al. (2016) \cite{casanova2016}. We also mention the best results achieved by the reviews of Mäenpää, T. and Pietikäinen, M. (2004) \cite{maenpaa2004} and Cernadas et al. (2017) \cite{cernadas2017}. It is important to notice that some of these works \cite{maenpaa2004,choice2011,cernadas2017} do not consider leave-one-out cross-validation, so the comparison should be taken cautiously.

%Gabor EEE, Hoang et al. (2002/2005) \cite{gabor2002,gabor2005};

We also considered the performance of methods based on four recent deep convolutional networks. The following pre-trained architectures are considered: AlexNet \cite{krizhevsky2012imagenet}; VGG-19 \cite{simonyan2014very}; GoogLeNet (Inception-v3) \cite{szegedy2016rethinking}; and ResNet50 (with 50 layers) \cite{he2016deep}. In this sense, as proposed in \cite{lin2013network}, each neural network was used as a feature extractor by applying the global average pooling (GAP) over the feature maps produced by its last convolutional layer. All the networks were imported from the PyTorch and Keras libraries. Since these models were previously trained with images of size $224 \times 224$ from the ImageNet dataset, images from the 5 datasets considered here were resized to $224 \times 224$ before being processed by these methods. Additionally, to show the improvement achieved by incorporating color into the CN approach, the previous gray-level CN method of Backes et al. (2013) \cite{backes} is also considered for comparison. All results are shown in Table \ref{comparison}, best results on each dataset are highlighted in bold type.

\begin{table}
	\caption{Classification accuracy rate of literature methods compared to the proposed method on 5 color-texture datasets. Empty cells indicates that the result for the correspondent method is not available for that dataset. The mean accuracy is only computed for methods with results on the 5 datasets.}
	\label{comparison}
	\centering
	\resizebox{\linewidth}{!}{
		\begin{tabular}{c|c|c|c|c|c|c}
			
			&Usptex&Vistex&Outex13&CURet& \ MBT \ &\\
			\hline
			\hline
			%\textit{Color-texture methods}&&&&\\
			Color statistics &64.7 & 85.2 & 76.4 & 39.3 &14.8&\\
			%Hoang et al. (2002/2005) \cite{gabor2002,gabor2005}&94.0&97.6&86.5&&\\    
			Mäenpää and Pietikäinen (2004) \cite{maenpaa2004}& & 99.5& 94.6&&&\\
			
			Qazi et al. (2011) \cite{choice2011} && 99.5& 94.4&&&\\
			
			Backes et al. (2012) \cite{backes2012} &96.6 & 99.1&&&&\\
			
			Sá Junior et al. (2014) \cite{sajunior2014} &96.9& 99.1&91.5&&&\\

			Casanova et al. (2016) \cite{casanova2016}& 97.0 &99.3 & 95.0&& & \\    
			
			Cernadas et al. (2017) \cite{cernadas2017}& 96.8& 97.3& 90.4& 95.5& &\\
			\hline 
			\textit{Deep Convolutional Networks}&&&&&&mean acc.\\
			AlexNet (2012) \cite{krizhevsky2012imagenet} & 98.5 & 99.4 & 87.5 & 92.8 & 90.54 & 93.7 ($\pm 5.1$) \\
			
			VGG-19 (2014)  \cite{simonyan2014very} & 97.6 & 97.9 & 88.3 & 94.0 & 91.07 & 93.8 ($\pm  4.2$) \\
			Inception-v3 (2016) \cite{szegedy2016rethinking} & 95.9 & 97.3 & 82.7 & 97.8& 93.79 &93.5 ($\pm 6.2$)\\
			ResNet50 (2016) \cite{he2016deep} & \textbf{99.8} & 99.7 & 91.7 & \textbf{98.5}& 95.74 &97.0 ($\pm 3.4$)\\
			
			%FV-SIFT FC+FV-VD \cite{cimpoi2016} &&&&99.7\\
			%ScatNet &&&&99.7\\
			\hline
			\textit{Complex Network methods}&&&&&\\
			Backes et al. (2013) \cite{backes} (gray-level)& 92.3 & 98.0 & 86.8 & 84.2& 83.7 &89.0 ($\pm 6.1$) \\
			%Backes et al., integrative (2013) \cite{backes}& 98.4 & 99.3 & 93.0 &  92.5& \textbf{98.3} &96.3 ($\pm 3.3$)\\
			\textbf{Multilayer CN descriptors }&99.0&\textbf{99.9}&\textbf{95.4}& 97.0& \textbf{97.1} &\textbf{97.7} ($\pm 1.8 $) \\
			
		\end{tabular}
	}
\end{table}

The baseline results computed with Color statistics gives a brief idea of the difficult to classify color-texture on each dataset. Vistex seems to be the most simple dataset, where Color statistics achieves 85.2\% of accuracy rate. This indicates that most of Vistex classes can be discriminated without considering pixels spatial interactions. Most of the literature color-texture descriptors (First part of Table \ref{comparison}) achieves results above 99\%, the gray-level CN method achieves 98.0\%, and the proposed method achieves the maximum accuracy rate of 99.9\%. The convolutional network ResNet50 achieves the second best result with 99.7\%. On the other hand, the convolutional networks VGG-19 and Inception-v3 achieve inferior results when compared to most of the other descriptors (97.9\% and 97.3\%, respectively).

On the Usptex dataset, the performance of Color statistics drops considerably, achieving 64.7\%. This dataset has a higher number of classes, therefore a simple first order analysis is not enough to provide acceptable results. Here we can notice that the performance of literature color-texture descriptors, around 96.6\% and 97.0\%, is similar to the convolutional networks AlexNet, VGG-19 and Inception-v3, varying around 95.9\% (Inception-v3) and 98.5\% (AlexNet). The lowest results are achieved using the gray-level CN method (92.3\%), suggesting that color help discriminates between the Usptex classes. The highest result is achieved with the ResNet50 convolutional network (99.8\%), followed by the proposed method (99.0\%).

The higher accuracy rates obtained on Vistex and Usptex difficult the comparison of performance between methods, for instance, results very close to 100\% as achieved by ResNet50 and the proposed method. Moreover, the CN gray-level method performs above 90\% on these datasets as well, which raise the question whether the performance gain from adding color is applicable. The other datasets give a better understanding of the differences between each method due to its higher difficulty of discrimination. On CURet, Color statistics perform poorly with an accuracy rate of 39.3\%, which corroborates to the difficulty of this dataset. The highest accuracy rates are obtained by ResNet50 with 98.5\%, Inception-v3 with 97.8\% and the proposed method with 97.0\%. %Considering the higher number of samples of each class on this dataset (92), we believe that this slight fluctuation of the proposed method performance happens due to overfitting and/or overtuning of parameters, or perhaps a higher number of features is needed.

On Outex13 and MBT, color is a key factor to the discrimination of texture classes because these datasets were built exclusively for color-texture evaluation. On these datasets, we can notice that the performance of all convolutional networks drops considerably. On Outex13, the methods AlexNet (87.5\%), VGG-19 (88.3\%) and Inception-v3 (82.7\%) performs similarly to the gray-level CN method, that achieves 86.8\%. The literature color-texture methods also overcome the performance of all convolutional networks, where the method of Casanova et al. \cite{casanova2016} reaches 95.0\%. The proposed method achieves the highest performance, with 95.4\%, which corroborates to the ability of Multilayer CN descriptors to deal with color-texture.

The last dataset analyzed, MBT, is perhaps the most challenging in terms of color-texture discrimination. Here, Color statistics can barely discriminate texture classes, achieving 14.8\% of accuracy rate. This should be expected, as the authors of MBT argue \cite{abdelmounaime2013newbrodatz} that the dataset was built so that no first-order statistic can discriminate between its classes. No results were found available for the literature color-texture methods considered on our comparisons. We then analyze the performance of the convolutional networks, the gray-level CN method, and the proposed method. The highest classification accuracy is obtained with the proposed method (97.1\%), and the lowest by the gray-level CN method (83.7\%). Convolutional networks performs around 90.54\% (AlexNet) and 95.74\% (ResNet50). In contrast to Outex13, it is possible to notice that the performance loss of these methods is related to the importance of color to the discrimination of texture.

%Similarly, color statistics cannot discriminate between the CURet classes, where the method achieves 39.3\% of accuracy rate. On Usptex and Outex13, the results (64.7\% and 76.4\%, respectively) of Color statistics are relatively higher

Concerning the gray-level CN method \cite{backes} and the proposed Multilayer CN method, significant improvement is achieved on all 5 datasets, except for Vistex where previous results were already near 100\%. The highest improvement is achieved on the MBT dataset, where the accuracy rate was improved from 83.7\% to 97.1\%. The overall performance over the 5 datasets (i.e. the mean accuracy rate) was improved from 89.0 ($\pm 6.1$) of the CN gray-level method to 97.7 ($\pm 1.8$) of the proposed method. Among the convolutional networks, ResNet50 achieves the highest overall performance, with (97.0 ($\pm 3.4$)), while other networks perform around 93\%. In comparison with the proposed method, ResNet50 has a lower overall performance and a higher standard deviation, highlighting its performance loss on the Outex13 and MBT datasets.

%In the Outex13 dataset the best results are achieved by \cite{maenpaa2004} and \cite{choice2011}. Our result in this dataset surpasses the graph-based method \cite{sajunior2014} and the method of Cernadas et al., which uses a technique invariant to changes in illumination color (GrayWorld Normalization) \cite{cernadas2017}.

% \begin{table}
% \caption{Comparisons with state-of-the-art methods on the PFID61 dataset. Results achieved using the configuration previously described, same as in \cite{farinella2014,martinel2015,zou2017} (3-fold cross-validation, taking 6 samples from 2 instances (12 images) for training and 6 samples from the another instance as test). }
% \centering
% \begin{tabular}{c|c}
%     &Accuracy rate (\%)\\
%     \hline   
%     Farinella et al. (2014) \cite{farinella2014} &31.3\\
%     Martinel et al. (2015) \cite{martinel2015} & 48.1\\
%     Zhou et al. (2017) \cite{zou2017} & 48.7\\
%     Proposed method & 28.6\\
%     \hline
%     \end{tabular}
% \end{table}

% \begin{table}
% \caption{Comparisons with state-of-the-art methods on the KTH-tips 2b dataset. Results achieved using the configuration previously described, same as in \cite{khan2015,sulc2016,cimpoi2016} (4-fold cross-validation, taking samples from 3 instances as training and from 1 instance for test).}
% \centering
% \begin{tabular}{c|c}
%     &Accuracy rate (\%)\\
%     \hline
%     Khan et al. (2015) \cite{khan2015} &70.6\\
%     Šulc and Matas (2016) \cite{sulc2016} &77.2 \\
%     Cimpoi et al. (2016) \cite{cimpoi2016} & 81.8\\

%     Proposed method & 54.0\\
%     \hline
%     \end{tabular}
% \end{table}

\section{Conclusion}\label{sec:conclusion}

In this work, we introduce a novel method for the modeling and characterization of color-texture motivated by CN and the within-between color-channel spatial interaction of pixels. Our method, called Multilayer CN descriptors, models the image as a multilayer CN with connections between distinct pixels from all color-channels of the image. The structure of these networks highlight the spatial interaction within-between color-channels that are quantified with traditional CN measures, composing a color-texture descriptor. Moreover, we also propose a new technique to the automatic selection of CN thresholds, an issue of previous works.

%It consists of computing the edge weight distribution $P(w)$ of CNs built for training images. We found that the peak of $P(w)$ can be set as the lower threshold limit $t_1$, which implies on discarding denser networks with mean degree around the maximum. On the other hand, the upper threshold limit $t_m$ is defined where $\int ^{t_m} P(w) dw = 0.9$, which then discard sparse networks with mean degree above 1. Then, our threshold set $T$ is defined by $m$ values between $[t_1, t_m]$, and we suggest the use of $m=10$.

%To validate the proposed method, our experiments consists of classification in 5 known color-texture datasets, named Usptex, Vistex, Outex13, CURet, and MBT. We found that the number of thresholds $m$ around $[8, 14]$ presents the best results, so we suggest the use of $m=10$. The combination of radii is more effective than using individual radius, and the best results are achieved using 5 or 6 radii.

Results indicate that the combination of within and between color-channel analysis provides rich discriminant features in most datasets, except for MBT where within-channel analysis proved to be more effective alone. The proposed method significantly improves the previous gray-level CN method \cite{backes} in all cases, where the overall performance (mean accuracy rate under the 5 datasets) of 89.0\%($\pm 6.1$) is improved to 97.7\%($\pm 1.8$). We then compare our results with several literature approaches, including traditional color-texture descriptors and the deep convolutional networks AlexNet, VGG-19, Inception-v3 and ResNet50 (with global average pooling). According to our findings, convolutional networks presents significant performance loss when color is a key property to discriminate texture classes (Outex13 and MBT datasets), performing below traditional color-texture methods on Outex13, for instance. The overall performance achieved by the best convolutional network, ResNet50, is 97.0\% ($\pm 3.4$). The use of multilayer CN proved to be more effective in this case, thus we believe that it should be further explored. As a future work, the analysis of the pertinent choice of different color spaces should be considered.

%Despite the great performance of convolutional networks on other tasks, the same is not observed for problems such as color-texture analysis. The spatial analysis of pixels and the within-between color-channel analysis stills provides interesting results for color-texture characterization without involving complex and exhausting training processes with enormous image sets.

\section*{Acknowledgments}

We would like to thank Abdelmounaime Safia for the feedback concerning the MBT dataset construction. L. F. S. Scabini acknowledges support from CNPq (Grant \#134558/2016-2). O. M. Bruno acknowledges support from CNPq (Grant \#307797/2014-7 and Grant \#484312/2013-8) and FAPESP (grant \#14/08026-1). R. H. M. Condori acknowledges support from Cienciactiva, an initiative of the National Council of Science, Technology and Technological Innovation-CONCYTEC (Peru). W. N. Gon\c{c}alves acknowledges support from CNPq (Grant \#304173/2016-9) and Fundect (Grant \#071/2015).

%
% ---- Bibliography ----
%
\bibliographystyle{IEEEtran}
\bibliography{refs}

\end{document}